\documentclass[lettersize,journal]{IEEEtran}

\usepackage{amsmath,amsfonts,amssymb}
\usepackage{array}
\usepackage{booktabs}
\usepackage{cite}
\usepackage{graphicx}
\usepackage{multirow}
\usepackage{pifont}
\usepackage{placeins}
\usepackage{stfloats}
\usepackage{textcomp}
\usepackage{url}
\usepackage[table]{xcolor}

\newcommand{\cmark}{\ding{51}}
\newcommand{\xmark}{\ding{55}}

\definecolor{highlightgray}{gray}{0.9}
\definecolor{highlightgray_light}{gray}{0.96}

\usepackage[hidelinks]{hyperref}

\title{GS$^{\mathbf{2}}$CI: Robust Gaussian Splatting For \\
Snapshot Compressive Imaging via Large Vision Model Priors}

\author{Yanming~Yang, Chenxi~Song, Ping~Wang, Xin~Yuan, and Chi~Zhang%
\thanks{\hspace*{-1em}Corresponding author: Chi Zhang (chizhang@westlake.edu.cn). \\ Yanming Yang, Chenxi Song and Chi Zhang are with the AGI Lab, Westlake University, Hangzhou 310030, China. Ping Wang and Xin Yuan are with Westlake University, Hangzhou 310030, China. (e-mail: \{yangyanming, songchenxi, wangping, xylab, chizhang\}@westlake.edu.cn).}}

\makeatletter
\let\GSorigmaketitle\maketitle
\let\GSorigatmaketitle\@maketitle
\let\GSorigthanks\thanks
\makeatother

\begin{document}

\maketitle

\begin{abstract}
Snapshot Compressive Imaging (SCI) offers an efficient solution for high-speed video acquisition and, under exposure-time camera--scene relative motion, multi-view scene capture by compressing temporal or spatial information into a single 2D measurement. While recent studies have explored SCI for 3D scene reconstruction, existing methods struggle with significant challenges due to information loss, limited viewpoint diversity, and the computational burden of jointly optimizing 3D representations and camera poses. In this work, we propose a novel framework that reconstructs high-quality 3D scenes from a single SCI measurement by leveraging 3D Gaussian Splatting (3DGS) and the powerful priors of large-scale vision foundation models (VFMs). Our primary reconstruction combines measurement-derived 3D VFM initialization with SCI-aware Gaussian optimization. After coarse-stage convergence, an auxiliary 2D VFM provides pseudo-view supervision at synthesized viewpoints for local appearance refinement. To further address the instability caused by ambiguous SCI supervision during 3DGS optimization, we introduce Opacity-Guided Splitting and Growth Regulation (OSGR), an SCI-specific densification strategy that augments split candidates using local opacity statistics, discourages loss-compensating opacity inflation through mean-opacity regulation, and bounds representation growth with explicit candidate-ratio and Gaussian-count constraints. Extensive experiments across multiple benchmarks demonstrate that our method achieves the strongest overall performance, combining leading reconstruction quality and robustness to viewpoint variation with competitive computational efficiency. Our code is available at {\def\UrlBreaks{\do\/\do-\do.}\url{https://github.com/Westlake-AGI-Lab/GS2CI.git}}.
\end{abstract}


\begin{figure*}[!t]
    \centering
    \includegraphics[width=\textwidth]{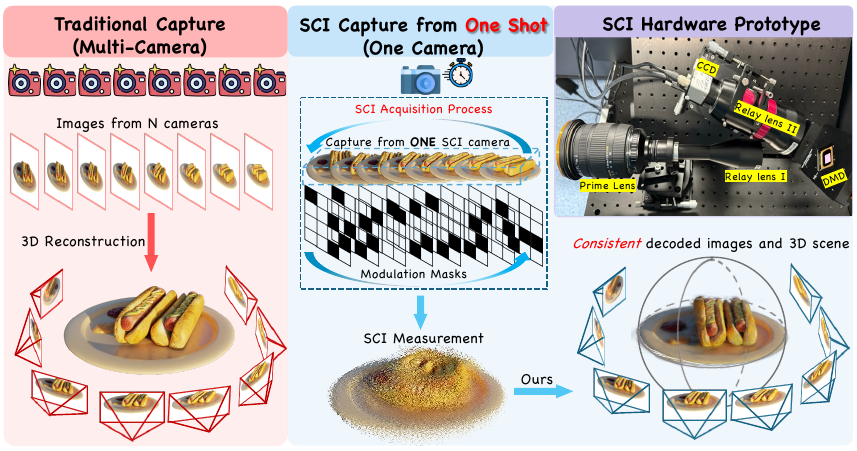}
    \caption{Conceptual comparison between conventional multi-camera capture and video SCI acquisition. Under rigid camera--scene relative motion, one SCI camera sequentially modulates views observed at different instants and accumulates them into a single sensor measurement. Our method reconstructs the shared static scene and the corresponding camera trajectory from this measurement and the coding masks.}
    \label{fig:teaser}
\end{figure*}

\section{Introduction}
\label{sec:intro}
\IEEEPARstart{S}{napshot} Compressive Imaging (SCI)~\cite{sci-rank} is a cutting-edge technique designed for high-speed data acquisition by compressing temporal information into a single 2D image. This is accomplished by modulating the incoming light using a sequence of specially designed 2D masks during the exposure process, allowing multiple temporal frames to be encoded into a single snapshot. A dedicated decoder is then required to reconstruct the original video frames from the compressed measurement. SCI provides a compact and efficient solution for capturing dynamic scenes, significantly reducing storage requirements and hardware complexity. These advantages make SCI attractive for high-speed video recording, where conventional high-speed cameras are costly and impose substantial data-throughput and storage demands.

Most existing approaches, whether model-based~\cite{sci-rank} or learning-based~\cite{admmnet,birnat,cheng2021memoryefficientnetworklargescalevideo,wang2022spatialtemporaltransformervideosnapshot,wang2024hierarchicalseparablevideotransformer,wang2023efficientscidenselyconnectednetwork,yuan2020plugandplayalgorithmslargescalesnapshot,yuan2021plugandplayalgorithmsvideosnapshot}, are therefore designed with a video-decoding perspective: they aim to recover a sequence of 2D frames under limited inter-frame viewpoint variation. As a result, they struggle to generalize when the SCI measurement encodes multi-view inputs with wide baseline disparities. This performance drop can be attributed to two factors: (i) the use of relatively small and low-diversity training datasets for SCI decoders, such as DAVIS~\cite{ponttuset20182017davischallengevideo}, and (ii) the lack of explicit modeling of essential 3D properties, including depth, occlusion, and view-dependent appearance. Consequently, although these decoders perform well on sequences with modest motion and viewpoint continuity, their effectiveness degrades under large viewpoint changes or discontinuous observations, leading to temporally and spatially inconsistent reconstructions.

During one sensor exposure, video SCI sequentially applies coding masks to temporally varying observations. When this variation is induced by rigid camera--scene relative motion, the latent frames correspond to different viewpoints and are multiplexed into one sensor readout. A single SCI camera can therefore encode multi-view information without simultaneous capture by a camera array, while the available viewpoint span is determined by the relative motion completed during the exposure. As illustrated in Fig.~\ref{fig:teaser}, this acquisition model provides geometric cues with simpler hardware, yet existing methods largely treat SCI as an extreme video compression scheme and do not fully exploit its underlying 3D structure.

These observations motivate a shift in perspective, from pure video reconstruction toward recovering a coherent underlying 3D scene that explains the SCI measurement. From this viewpoint, SCI can be formulated as a joint problem of compressed image decoding and 3D scene recovery, where explicit 3D priors act as consistency constraints across views or time. Recent methods follow this direction by adopting implicit or explicit 3D representations, such as Neural Radiance Fields (NeRF)~\cite{mildenhall2020nerfrepresentingscenesneural} and 3D Gaussian Splatting (3DGS)~\cite{kerbl20233dgaussiansplattingrealtime}, to recover spatially consistent scenes from severely underdetermined SCI inputs~\cite{wang2024scigs3dgaussianssplatting,li2024scinerfneuralradiancefields,li2024learningradiancefieldssingle,sci-gaussian}.

However, because SCI compresses multiple viewpoints through coded modulation and pixel-wise aggregation, it inevitably discards fine-grained view cues, making geometry and pose estimation particularly ill-posed. Existing 3D-SCI methods typically rely on jointly optimizing 3D scene representations and camera parameters, which is computationally intensive and sensitive to initialization. For example, NeRF-based approaches~\cite{li2024scinerfneuralradiancefields} often require tens of thousands of iterations per scene with costly gradient computations. When only a single SCI measurement is available, the optimization tends to overfit the modulated views and generalizes poorly to unseen viewpoints, undermining both the decoding of compressed images and the recovery of a stable 3D scene without strong priors or external guidance.

To overcome these limitations, we present a computationally efficient framework built on 3DGS~\cite{kerbl20233dgaussiansplattingrealtime}, enhanced by vision foundation models trained on large-scale multi-view and 3D data. In our formulation, these foundation models provide strong priors on geometry, appearance, and camera motion, thereby regularizing the severely under-constrained SCI reconstruction problem. Our framework comprises a primary reconstruction stage and an auxiliary refinement stage. In the primary stage, a 3D VFM~\cite{wang2025vggtvisualgeometrygrounded} uses measurement-derived proxy views to initialize camera poses and sparse scene geometry, followed by SCI-aware Gaussian optimization. After coarse-stage convergence, the auxiliary stage uses a frozen 2D VFM to construct pseudo-view targets at synthesized poses for local appearance refinement.

While 3DGS offers clear advantages in rendering speed and training efficiency, applying it directly to SCI remains challenging because a single multiplexed measurement constrains only the encoded sum of the rendered views rather than each latent view independently. Different combinations of geometry, appearance, and opacity can therefore produce similar measurement values. During joint scene-and-pose optimization, the optimizer may reduce part of the multiplexed residual by increasing the opacity of individual Gaussians before their spatial support is sufficiently resolved, producing local opacity peaks that are not explicitly addressed by the original adaptive density control mechanism~\cite{kerbl20233dgaussiansplattingrealtime}.

To address this issue, we propose Opacity-Guided Splitting and Growth Regulation (OSGR), a densification strategy tailored to SCI. OSGR uses local opacity contrast to augment the split-candidate set, allowing concentrated responses to be reparameterized by smaller child Gaussians with distinct spatial support. The mean-opacity term softly discourages renewed opacity inflation, while the candidate-ratio and Gaussian-count constraints bound representation growth. Together, they guide refinement and stabilize densification under SCI supervision.

Extensive experiments on multiple benchmarks show that our method achieves the strongest overall performance by combining leading visual fidelity and novel-view synthesis quality with competitive reconstruction speed. Moreover, it demonstrates strong robustness to large viewpoint changes and challenging SCI settings. The main contributions of this paper are summarized as follows:
\begin{itemize}
    \item We propose a 3D reconstruction framework from a single SCI measurement that combines measurement-derived 3D VFM geometry initialization with SCI-aware Gaussian optimization, followed by auxiliary 2D VFM pseudo-view supervision for local appearance refinement.
    \item We develop OSGR, a customized densification strategy for 3D reconstruction from SCI, which stabilizes the joint optimization of camera poses and scene structure under severely under-constrained SCI supervision.
    \item Extensive experiments across multiple benchmarks demonstrate the strongest overall performance, with leading reconstruction quality and competitive efficiency. The code is publicly available.
\end{itemize}

\section{Related Work}
\label{sec:related-work}
\paragraph{Video SCI Reconstruction}
SCI aims to recover a video sequence from a single compressed measurement, significantly reducing acquisition cost and latency~\cite{sci-theory}. Due to its ill-posed nature, strong priors are required to constrain the solution space. Traditional methods adopt iterative optimization with handcrafted regularizers~\cite{shearlet,yuan2015generalizedalternatingprojectionbased,sci-rank}, but often suffer from slow convergence and poor scalability to high-resolution or dynamic scenes. Recent work has shifted to learning-based methods that directly map compressed measurements to videos, leveraging neural networks~\cite{hybrid,birnat,vaswani2017attention} to better model spatiotemporal dependencies. However, most learning-based methods rely on synthetic data and neglect 3D scene geometry, leading to geometric inconsistencies across views.

\paragraph{Geometric and Generative Priors}
Recent 3D reconstruction methods can be broadly grouped into two directions. The first direction explores 3D VFMs, which learn to infer depth, camera poses, and scene geometry from large-scale image data without explicit SfM supervision~\cite{wang2023visualgeometrygroundeddeep,zhang2024monst3rsimpleapproachestimating,jang2025pow3rempoweringunconstrained3d,yang2025fast3r3dreconstruction1000,li2025megasam,wang2024dust3r,wang2025pi3scalablepermutationequivariantvisual}. A representative example is VGGT~\cite{wang2025vggtvisualgeometrygrounded}, which jointly predicts pose and geometry in a feed-forward end-to-end framework. The second direction leverages generative priors, particularly diffusion models, to facilitate sparse-view reconstruction~\cite{wu2025difix3d+,xu2025breaking,yin2025gsfixer,wang2025fixinggs,wang2025videoscene}. These methods refine imperfect renderings with pretrained or distilled diffusion models and update the 3D representation, improving reconstruction quality and view consistency.

\section{Method}
We present a framework for SCI-based 3D reconstruction by integrating VFMs with a 3DGS-based SCI reconstruction pipeline. The primary reconstruction uses a 3D VFM~\cite{wang2025vggtvisualgeometrygrounded} to initialize geometry and camera poses before SCI-aware Gaussian optimization. An auxiliary fine stage subsequently applies 2D VFM pseudo-view supervision at synthesized viewpoints for local appearance refinement while keeping the camera trajectory estimated by the primary reconstruction fixed.

\begin{figure*}[!t]
    \centering
    \begingroup
    \setlength{\unitlength}{1pt}%
    \definecolor{proxyblockbg}{RGB}{223,239,243}%
    \resizebox{\linewidth}{!}{%
    \begin{picture}(425.055,227.141)
        \put(0,0){\includegraphics[width=425.055pt]{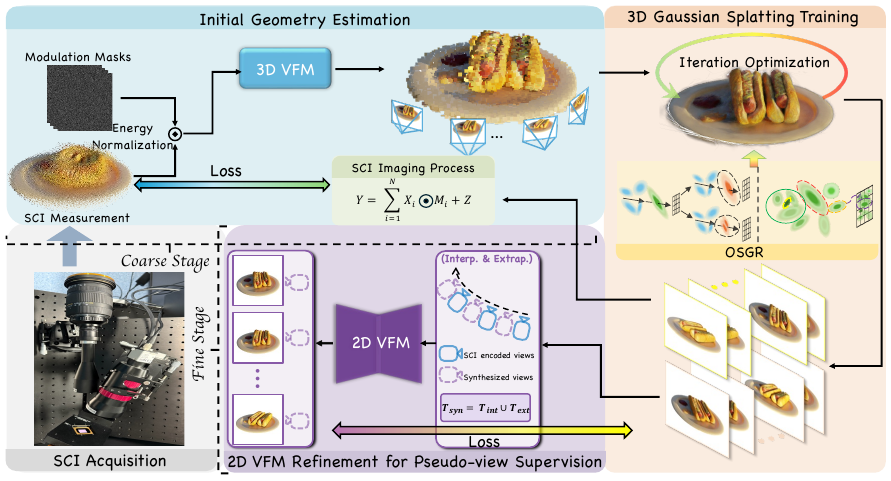}}
        \put(40.0,159.0){%
            \setlength{\fboxsep}{0pt}%
            \colorbox{proxyblockbg}{%
                \parbox[c][24pt][c]{44pt}{\centering\fontsize{5.3}{5.7}\selectfont\ttfamily Proxy\\Construction}%
            }%
        }
    \end{picture}%
    }
    \endgroup
    \caption{Pipeline overview of our 3D reconstruction framework from a single SCI measurement. Measurement-derived proxy views are processed by a 3D VFM~\cite{wang2025vggtvisualgeometrygrounded} to initialize scene geometry and camera poses, followed by SCI-aware 3DGS~\cite{kerbl20233dgaussiansplattingrealtime} optimization augmented by our OSGR. After coarse-stage convergence, an auxiliary fine stage uses a frozen 2D VFM to construct pseudo-view targets at synthesized poses for local appearance refinement.}
    \label{fig:pipeline}
\end{figure*}

The remainder of this section is organized as follows. We begin by introducing the preliminaries of our method in Section~\ref{sec:preliminaries}. We then describe the initial geometry and pose estimation using the 3D VFM in Section~\ref{sec:method-vggt}. The primary SCI reconstruction and auxiliary pseudo-view refinement are subsequently presented in Section~\ref{sec:method-framework} and Section~\ref{sec:method-repair}, respectively, followed by our improved densification strategy tailored for SCI in Section~\ref{sec:densification}. An overview of the entire framework is illustrated in Fig.~\ref{fig:pipeline}.

\subsection{Preliminaries} \label{sec:preliminaries}
\paragraph{Snapshot Compressive Imaging Model} 
In a typical SCI~\cite{sci-theory} system, the compressed image $ Y $ is formed by modulating the scene through $ N $ binary masks $ M_i $ during the exposure time. The sensor records the accumulated exposure over these masks. The image formation process can be mathematically represented as:
\begin{equation}
    \mathbf{Y} = \sum_{i=1}^{N} \mathbf{X}_i \odot \mathbf{M}_i + \mathbf{Z},
    \label{eq:sci}
\end{equation}
where $ Y $ is the compressed snapshot, $ X_i $ represents the $ i $-th modulated image corresponding to each exposure interval, $ M_i $ is the binary modulation mask, $ Z $ is the measurement noise and $\odot$ denotes Hadamard product. The number of masks $ N $ defines the temporal compression ratio (CR), which determines how many frames are compressed into a single snapshot. The pixel values in the binary mask are typically randomized or designed with a fixed overlap ratio. For the static-scene setting considered in this work, $\mathbf{X}_i$ denotes the scene observed at camera pose $\mathbf{T}_i$ during the $i$-th mask interval. Multi-view parallax is produced by camera--scene relative motion within the sensor exposure; the sequentially modulated observations are accumulated into one sensor readout. If the relative pose remains fixed, the latent frames share the same viewpoint and the measurement contains no multi-view parallax.

\subsection{Initial Geometry Estimation}
\label{sec:method-vggt}
A single SCI measurement $\mathbf{Y}$ lacks the view-resolved RGB observations required by conventional Structure-from-Motion methods such as COLMAP~\cite{colmap} and by VGGT~\cite{wang2025vggtvisualgeometrygrounded}. We therefore construct geometry-oriented proxy views on the original image grid from $\mathbf{Y}$ and the actual coding masks $\mathcal{M}=\{\mathbf{M}_i\}_{i=1}^{N}$. Let $a(\mathbf{p})=\sum_i\mathbf{M}_i(\mathbf{p})$ and $\mu_{\mathcal{M}}=(HW)^{-1}\sum_{\mathbf{p}}a(\mathbf{p})$ denote the per-pixel and mean measurement multiplicities, respectively.

For standard low-multiplicity acquisitions ($\mu_{\mathcal{M}}<\tau_{\mu}$, with $\tau_{\mu}=4$ fixed throughout), we use Energy-Normalized Initialization (ENI) to compensate for first-order spatial exposure variation~\cite{birnat} before assigning the normalized, mask-supported samples to individual proxy views:
\begin{equation}
\begin{gathered}
    \overline{\mathbf{Y}}(\mathbf{p})=
    \begin{cases}
        \mathbf{Y}(\mathbf{p})/a(\mathbf{p}), & a(\mathbf{p})>0,\\
        \mathbf{0}, & a(\mathbf{p})=0,
    \end{cases}\\
    \widetilde{\mathbf{X}}^{\mathrm{ENI}}_i
    =\mathcal{I}\!\left(
        \overline{\mathbf{Y}}\odot\mathbf{M}_i;\mathbf{M}_i
    \right),\qquad i=1,\ldots,N.
\end{gathered}
\label{eq:proxy-construction}
\end{equation}
Here, $\mathcal{I}$ completes each proxy by retaining the normalized measurement samples selected by $\mathbf{M}_i$ and interpolating the locations omitted by that mask. Linear interpolation is used inside the convex hull of the selected samples, while the remaining locations are filled by nearest-neighbor interpolation.

However, when increased compression or mask overlap yields $\mu_{\mathcal{M}}\geq\tau_{\mu}$, ENI cannot separate the superposed view contributions. We therefore use Adaptive Mask-Decoding Initialization (AMDI), which estimates the $N$ proxy views from local coding-pattern diversity through adaptive, regularized least squares. Both constructions provide geometry-oriented VGGT inputs rather than recovered frames or photometric supervision. The supplementary material details how measurement multiplicity determines the routing rule and provides the complete AMDI solver. Writing the two constructions as $\mathcal{P}_{\mathrm{ENI}}$ and $\mathcal{P}_{\mathrm{AMDI}}$, we define the proxy sequence as:
\begin{equation}
    \widetilde{\mathcal{X}}
    =
    \begin{cases}
        \mathcal{P}_{\mathrm{ENI}}(\mathbf{Y},\mathcal{M}),
        & \mu_{\mathcal{M}}<\tau_{\mu},\\
        \mathcal{P}_{\mathrm{AMDI}}(\mathbf{Y},\mathcal{M}),
        & \mu_{\mathcal{M}}\geq\tau_{\mu}.
    \end{cases}
    \label{eq:proxy-routing}
\end{equation}

\begin{samepage}
The resulting proxy sequence is supplied to the frozen VGGT model to estimate a 3D point set and camera parameters. Bundle adjustment (BA) is preferred when the proxy-derived correspondences provide reliable constraints, because it jointly refines the 3D structure and camera parameters to improve global multi-view consistency. Since these indirect correspondences can be spatially uneven, we verify the post-BA geometry before using it to initialize 3DGS and retain a supported estimate when BA is underconstrained. We denote this complete VGGT-based initialization as follows:
\begin{equation}
\left(
    \mathcal{Q}^{\star},
    \{\mathbf{K}^{\star}_i,\mathbf{T}^{(0)}_i\}_{i=1}^{N}
\right)
    =\Phi_{\mathrm{init}}\!\left(
        \widetilde{\mathcal{X}}
    \right).
    \label{eq:vggt}
\end{equation}
Here, $\mathcal{Q}^{\star}$ is the selected 3D point set, $\mathbf{K}^{\star}_i$ is the camera intrinsic matrix, and $\mathbf{T}^{(0)}_i$ is the initial camera pose for view $i$.
The reliability criteria are fixed a priori and applied consistently across all scenes. The treatment of the geometry before and after refinement, together with the handling of degenerate initializations, is described in the supplementary material.
\end{samepage}

\subsection{Pose Refinement and Scene Optimization}
\label{sec:method-framework}
After obtaining the initial point cloud from the 3D VFM~\cite{wang2025vggtvisualgeometrygrounded}, we jointly optimize the Gaussian scene and camera poses. Although the 3D VFM provides a strong initialization, small pose errors can hinder precise reconstruction. We therefore use a hybrid objective that enforces fidelity to the observed SCI measurement and penalizes mean Gaussian opacity. With~$\widehat{\mathbf{Y}}=R(\mathcal{G},\mathbf{T})$ denoting the rendered SCI measurement, the coarse loss is:
\begin{equation}
\begin{aligned}
    \mathcal{L}_{\mathrm{coarse}}={}&
    (1-\lambda_{\mathrm{ssim}})\mathcal{L}_1(\widehat{\mathbf{Y}},\mathbf{Y})
    +\lambda_{\mathrm{ssim}}\mathcal{L}_{\mathrm{ssim}}(\widehat{\mathbf{Y}},\mathbf{Y}) \\
    &+\frac{\lambda_{\mathrm{opacity}}}{G}\sum_{i=1}^{G}\alpha_i,
\end{aligned}
    \label{eq:coarse-loss}
\end{equation}
Here, \(R(\mathcal{G}, \mathbf{T})\) denotes the entire SCI imaging pipeline: the current Gaussians \(\mathcal{G}\) are rendered using poses \(\mathbf{T}\), and the rendered views are then encoded according to Eq.~\eqref{eq:sci}. Moreover, $\alpha_i$ denotes the opacity of the $i$-th Gaussian and $G$ is the number of Gaussians at the current training step.

Each initial pose \( \mathbf{T}_i^{(0)} \in \mathrm{SE}(3) \) is refined via a learnable adjustment \( \Delta\mathbf{T}_i \in \mathrm{SE}(3) \), yielding the final pose: $\mathbf{T}_i = \Delta\mathbf{T}_i \cdot \mathbf{T}_i^{(0)}$, where \( \Delta\mathbf{T}_i \) is optimized in the Lie algebra \( \mathfrak{se}(3) \).
Each adjustment is initialized as the identity transformation and jointly optimized with the Gaussian parameters under the coarse-stage objective in Eq.~\eqref{eq:coarse-loss}.

\subsection{Auxiliary 2D VFM Pseudo-view Refinement}
\label{sec:method-repair}

After the primary coarse-stage reconstruction, we first render the current Gaussian representation at synthesized viewpoints and process each rendering with a frozen 2D VFM~\cite{wu2025difix3d+} to produce a pseudo-ground-truth appearance target for the corresponding viewpoint. Because these targets are generated independently by a 2D model, they may contain local appearance content that is inconsistent with the current 3D scene representation. To reduce the influence of such inconsistencies during optimization, we assign a small overall weight to the pseudo-view objective and modulate its photometric residuals using a soft support weight derived from Gaussian alpha accumulation. With the recovered camera trajectory fixed and the SCI consistency objective retained, this auxiliary supervision further refines the local appearance of the Gaussian representation.

We synthesize target poses from the refined camera trajectory $\{\mathbf{T}_i\}_{i=1}^{N}$. Given two adjacent anchor poses $\mathbf{T}_a=(\mathbf{R}_a,\mathbf{t}_a)$ and $\mathbf{T}_b=(\mathbf{R}_b,\mathbf{t}_b)$, we interpolate their rotations on $\mathrm{SO}(3)$ and their translations linearly:
\begin{equation}
\begin{aligned}
    \mathbf{R}(\alpha)
    &=
    \exp\!\left(
    \alpha\log\!\left(\mathbf{R}_b\mathbf{R}_a^{\mathsf{T}}\right)
    \right)\mathbf{R}_a,\\
    \mathbf{t}(\alpha)
    &=
    (1-\alpha)\mathbf{t}_a+\alpha\mathbf{t}_b,\\
    \mathbf{T}_{\mathrm{int}}(\alpha)
    &=
    \begin{bmatrix}
        \mathbf{R}(\alpha) & \mathbf{t}(\alpha)\\
        \mathbf{0}^{\mathsf{T}} & 1
    \end{bmatrix}.
\end{aligned}
    \label{eq:pose_interp}
\end{equation}
where $\alpha\in(0,1)$. To generate the endpoint extrapolated poses, we extend the boundary motion by one additional step: the relative transform from the second pose to the first is applied once more before the first pose, and the relative transform from the penultimate pose to the last is applied once more after the last pose. The synthesized pose set is: \mbox{$\mathcal{T}_{\mathrm{syn}}=\mathcal{T}_{\mathrm{int}}\cup\mathcal{T}_{\mathrm{ext}}$}.

At each synthesized pose $\mathbf{T}_t \in \mathcal{T}_{\mathrm{syn}}$, the auxiliary renderer produces a pseudo-view and alpha map: \mbox{$(\mathbf{X}^{0}_t,\mathbf{A}^{0}_t)=\mathcal{R}_{\mathrm{aux}}(\mathcal{G},\mathbf{T}_t)$}. After coarse-stage optimization is completed, we apply the frozen 2D VFM $\mathcal{F}_{\psi}$ to obtain a refinement target: \mbox{$\hat{\mathbf{X}}_t=\mathcal{F}_{\psi}(\mathbf{X}^{0}_t;\tau_0)$}, where $\tau_0$ controls the refinement strength.

To construct the soft support weight, we additionally render the current fine-stage Gaussian representation at each synthesized pose as: \mbox{$(\mathbf{X}^{r}_t,\mathbf{A}^{r}_t)=\mathcal{R}_{\mathrm{aux}}(\mathcal{G},\mathbf{T}_t)$}. We compute an alpha-based support score as: \mbox{$\mathbf{A}^{\mathrm{sup}}_t=\min(\mathbf{A}^{0}_t,\mathbf{A}^{r}_t)$}, where the pointwise minimum retains only the accumulated alpha supported both when the pseudo-ground-truth target is generated and at the current optimization step. We then map this score to a soft pixel-level weight:
\begin{equation}
    \mathbf{W}_t(\mathbf{p})
    =
    w_{\min}
    +
    (1-w_{\min})
    \sigma\!\left(
    \frac{
        \mathbf{A}^{\mathrm{sup}}_t(\mathbf{p})-\tau_v
    }{\beta_v}
    \right),
    \label{eq:visibility_weight}
\end{equation}
where $\sigma(\cdot)$ is the sigmoid function, $\tau_v$ is the support threshold, $\beta_v$ controls the transition softness, and $w_{\min}$ retains a residual contribution for pixels with low accumulated alpha. The resulting weight approaches one for high-support pixels and $w_{\min}$ for low-support pixels. The corresponding hyperparameters are specified in Sec.~IV-A. We apply this lightweight weighting only to the photometric component of the auxiliary pseudo-view loss, while retaining a weak unweighted perceptual term. Let $d_t(\mathbf{p})=\ell_1(\mathbf{X}^{r}_t(\mathbf{p}),\hat{\mathbf{X}}_t(\mathbf{p}))$ denote the photometric error at pixel $\mathbf{p}$:
\begin{equation}
    \mathcal{L}_{\mathrm{syn}}^t
    =
    \lambda_{l1}
    \frac{
    \sum_{\mathbf{p}}
    \mathbf{W}_t(\mathbf{p})
    d_t(\mathbf{p})
    }{
    \sum_{\mathbf{p}}\mathbf{W}_t(\mathbf{p})+\epsilon
    }
    +
    \lambda_{\mathrm{per}}
    \mathcal{D}_{\mathrm{per}}
    \left(\mathbf{X}^{r}_t,\hat{\mathbf{X}}_t\right),
    \label{eq:visibility_repair_loss}
\end{equation}
where $\mathcal{D}_{\mathrm{per}}$ denotes a perceptual distance and $\epsilon$ is a small constant for numerical stability. This weighting reduces the contribution of photometric residuals along rays with low accumulated alpha.

The final fine-stage objective is defined as:
\begin{equation}
    \mathcal{L}_{\mathrm{fine}}
    =
    \mathcal{L}_{\mathrm{coarse}}
    +
    \lambda_{\mathrm{syn}}
    \frac{1}{|\mathcal{T}_{\mathrm{syn}}|}
    \sum_{\mathbf{T}_t \in \mathcal{T}_{\mathrm{syn}}}
    \mathcal{L}_{\mathrm{syn}}^t,
    \label{eq:final_loss}
\end{equation}
where $\lambda_{\mathrm{syn}}$ controls the contribution of the auxiliary supervision at both interpolated and extrapolated pseudo-view poses.

\subsection{OSGR: Opacity-Guided Splitting and Growth Regulation}
\label{sec:densification}

In the original 3DGS framework~\cite{kerbl20233dgaussiansplattingrealtime}, adaptive density control (ADC) relies on \textit{duplicate} and \textit{split} operations to refine the Gaussian representation. While effective under standard multi-view supervision, these densification strategies become unstable in the SCI setting, where multiple views are compressed into a single measurement and the supervision is therefore severely under-constrained. In practice, standard ADC may densify weakly observed regions, causing abrupt opacity increases, photometric loss oscillations, and unstable joint scene-and-pose optimization.

To address this issue, we propose OSGR, a densification strategy tailored to SCI. OSGR incorporates local opacity statistics into candidate selection and regulates the global opacity distribution during optimization. It combines an opacity-guided splitting rule, a mean-opacity regulation term, and explicit candidate-ratio and Gaussian-count constraints.

\begin{figure*}[!t]
    \centering
    \includegraphics[width=\textwidth]{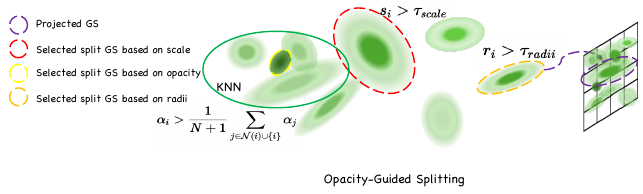}
    \caption{Illustration of the proposed OSGR strategy. In addition to the scale- and screen-radius criteria in vanilla 3DGS ADC~\cite{kerbl20233dgaussiansplattingrealtime}, OSGR identifies opacity-inconsistent Gaussians using local neighborhood statistics and applies global opacity regulation during optimization. Best viewed by zooming in.}
    \label{fig:enter-label}
\end{figure*}

Because a single SCI measurement constrains only the encoded sum of the rendered views, different combinations of geometry, appearance, and opacity can produce similar measurement values. The optimizer may consequently absorb part of the multiplexed residual by increasing the opacity of individual Gaussians, producing local opacity peaks during joint scene-and-pose optimization. OSGR uses these peaks as an additional signal for spatial refinement. For each Gaussian \(i\), we compute the average opacity within its local neighborhood:
\begin{equation}
    \bar{\alpha}_i = \frac{1}{K + 1} \sum_{j \in \mathcal{K}(i) \cup \{i\}} \alpha_j,
    \label{eq:opacity-average}
\end{equation}
where \(\mathcal{K}(i)\) denotes the neighbors of the \(i\)-th Gaussian. Building upon the default split mechanism in standard ADC, a Gaussian with $\alpha_i>\bar{\alpha}_i$ is included in the additional split set. Splitting converts the locally concentrated opacity response into smaller child Gaussians with distinct spatial support. Their different projections across views and masks allow subsequent SCI optimization to redistribute the contribution among the children, while the child-opacity correction limits the immediate compositing change caused by the split. Mean-opacity regulation discourages renewed opacity inflation, and the candidate-ratio and Gaussian-count constraints bound representation growth. The original ADC candidates remain active and are augmented by the opacity-guided candidates, as illustrated in Fig.~\ref{fig:enter-label}.

In addition to the local splitting rule, the global mean-opacity term discourages representation-wide opacity inflation during optimization, thereby providing a soft constraint on opacity values. The candidate-ratio budget limits additional candidates at each densification step, while the Gaussian-count constraint sets a hard cap on the total population.

Together, local opacity guides additional refinement, mean-opacity regulation constrains opacity evolution, and the explicit growth constraints bound representation size, stabilizing densification under SCI supervision.

\section{Experiment}
To evaluate the effectiveness and generalizability of the proposed method, we conduct comprehensive experiments under diverse SCI settings on synthetic datasets and real data, covering variations in scene type, camera motion, and novel view synthesis scenarios.

\subsection{Experimental Settings}
\paragraph{Datasets and Metrics}
To ensure comprehensive evaluation, we test our method on both simulated and real-world SCI datasets. Following prior work~\cite{li2024scinerfneuralradiancefields}, we first evaluate on simulated SCI measurements generated from widely used benchmarks: NeRF Synthetic~\cite{mildenhall2020nerfrepresentingscenesneural}, DeblurNeRF~\cite{ma2022deblurnerfneuralradiancefields}, DTU~\cite{jensen2014large}, LLFF~\cite{mildenhall2019llff} and DAVIS~\cite{ponttuset20182017davischallengevideo}. Unless otherwise specified, the simulated measurements use a compression ratio of 8 and a mask ratio of 0.25. For real-world evaluation, we use the real SCI data provided by SCINeRF~\cite{li2024scinerfneuralradiancefields}. We report SSIM, PSNR, and LPIPS as our main metrics. 

\paragraph{Compared Baselines}
We evaluate our method against a wide range of state-of-the-art SCI reconstruction approaches, including GAP-TV~\cite{yuan2015generalizedalternatingprojectionbased}, PnP-FFDNet~\cite{yuan2020plugandplayalgorithmslargescalesnapshot}, PnP-FastDVDNet~\cite{yuan2021plugandplayalgorithmsvideosnapshot}, and EfficientSCI~\cite{wang2023efficientscidenselyconnectednetwork}. We also compare with 3D-based state-of-the-art methods including SCINeRF~\cite{li2024scinerfneuralradiancefields} and SCIGS~\cite{wang2024scigs3dgaussianssplatting}. All experiments are evaluated under consistent settings for fair comparison.

\paragraph{Implementation Details}
We implement our method in Nerfstudio~\cite{Tancik_2023} and use gsplat for differentiable 3DGS rasterization~\cite{kerbl20233dgaussiansplattingrealtime}. All experiments use one NVIDIA RTX 4090 with a fixed random seed of 42, and the hyperparameters remain unchanged across standard and extended evaluations. On this GPU, proxy construction and the complete VGGT-based geometry initialization take 4 minutes 45 seconds on average. Coarse training takes 53.67 minutes, while DiFix3D+ target generation and subsequent fine-stage optimization add roughly 10 minutes, bringing the total to approximately 68.4 minutes. A fresh SCINeRF~\cite{li2024scinerfneuralradiancefields} run under its official schedule requires 746.84 minutes (12.45 hours).

\paragraph{Optimization Protocol}
\emph{Initialization and coarse stage.} We apply the acquisition-aware proxy construction and VGGT-based geometry initialization described in Section~\ref{sec:method-vggt} to every setting. The AMDI solver and the treatment of geometrically degenerate initializations are detailed in the supplementary material. The Gaussian representation and per-view local $\mathrm{SE}(3)$ pose corrections are then jointly optimized with Adam for 20,000 iterations. The SCI consistency objective combines $\ell_1$ and SSIM terms with weights of 0.9 and 0.1, while opacity regulation is weighted by 0.01. Pose gradients are accumulated over 25 iterations to stabilize camera refinement. The learning rates for pose corrections and Gaussian means decay exponentially from $10^{-2}$ and $1.6 \times 10^{-4}$ to $2 \times 10^{-5}$ and $1.6 \times 10^{-6}$, respectively.

\emph{Densification.} Densification is performed every 100 iterations between iterations 500 and 15,000. To constrain model growth, the Gaussian count at iteration 7,000 bounds all subsequent splitting and duplication, while opacity is reset every 3,000 iterations. OSGR augments the conventional gradient, scale, and screen-space radius cues with the proposed local opacity criterion. For duplication, opacity is revised following Revising-GS~\cite{bulo2024revisingdensificationgaussiansplatting} to preserve transmittance; for splitting, we extend the same transmittance-preserving correction to the resulting child Gaussians.

\emph{Auxiliary fine stage.} The fine stage is initialized from the final coarse-stage estimate and refines the Gaussian representation for 3,000 iterations with fixed camera poses. DiFix3D+~\cite{wu2025difix3d+} serves as the 2D VFM, with pseudo-view targets obtained through single-step inference at diffusion timestep $t=400$. Both pseudo-view groups use an outer loss weight of 0.006 and combine $\ell_1$ and perceptual terms with weights of 0.8 and 0.01. The alpha-based support weight is computed from the pointwise minimum of the target-construction and current-render alpha maps, with a support threshold of 0.25, transition softness of 0.08, and minimum weight of 0.10. For Table~\ref{tab:ablation_fine_stage}, the evaluation poses are disjoint from those used for pseudo-view supervision.

\subsection{Results}
\begin{figure*}[!t]
  \centering
  \includegraphics[width=\linewidth]{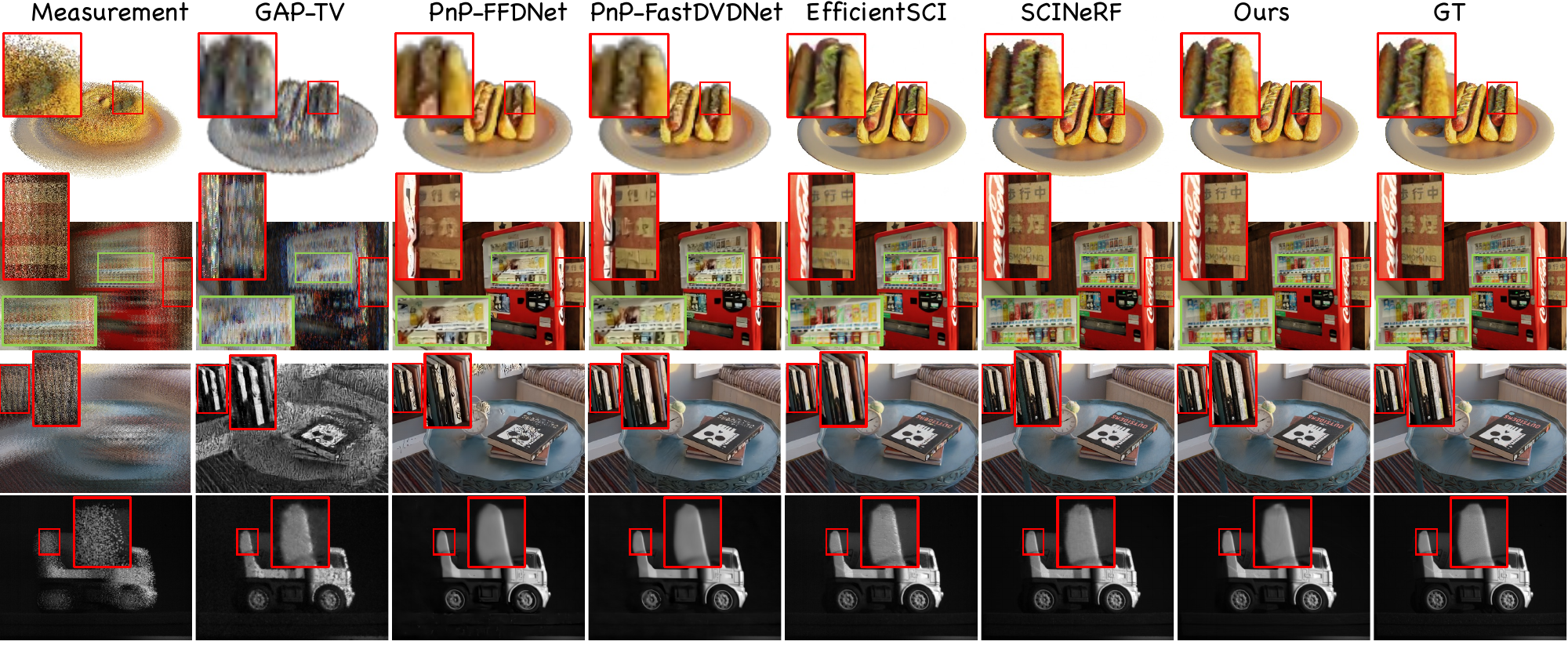}
  \caption{Qualitative evaluations of our method against state-of-the-art SCI decoding methods. Our results demonstrate the effectiveness of the proposed approach in restoring high-quality images from a single SCI measurement. The last row uses real SCI data provided by SCINeRF~\cite{li2024scinerfneuralradiancefields}.}
  \label{fig:sci-rec}
\end{figure*}

\paragraph{SCI Reconstruction}
Table~\ref{tab:sci-comparison} presents the quantitative comparison. Compared with EfficientSCI~\cite{wang2023efficientscidenselyconnectednetwork}, the state-of-the-art method for video sequence SCI, our approach achieves better multi-view consistency and higher reconstruction quality. In comparison to SCINeRF~\cite{li2024scinerfneuralradiancefields} and SCIGS~\cite{wang2024scigs3dgaussianssplatting}, which represent the state-of-the-art in 3D modeling-based SCI, our method achieves better overall performance across most scenes. These results confirm the effectiveness of the proposed method for reconstructing high-quality 3D scenes from SCI measurements. Visual comparisons are shown in Fig.~\ref{fig:sci-rec}.

\begin{table*}[t]
\caption{Quantitative Comparison of SCI Reconstruction Quality on Six Scenes. We report PSNR$\uparrow$, SSIM$\uparrow$, and LPIPS$\downarrow$. Our method achieves the best or second-best performance on almost all scenes and metrics, with particularly notable gains in perceptual quality (LPIPS) and structural fidelity (PSNR/SSIM), demonstrating its effectiveness for high-fidelity SCI reconstruction. \textbf{Bold} and \underline{underlined} values denote the best and second-best results, respectively.}
\label{tab:sci-comparison}
\centering
\small
\resizebox{\textwidth}{!}{%
\begin{tabular}{@{}l*{18}{@{\hspace{1pt}}c}@{}}
\toprule
Method &
\multicolumn{3}{c}{Vender} &
\multicolumn{3}{c}{Factory} &
\multicolumn{3}{c}{Airplants} &
\multicolumn{3}{c}{Hotdog} &
\multicolumn{3}{c}{Tanabata} &
\multicolumn{3}{c}{Cozy2room} \\
& ~PSNR & ~SSIM & ~LPIPS 
& ~PSNR & ~SSIM & ~LPIPS 
& ~PSNR & ~SSIM & ~LPIPS 
& ~PSNR & ~SSIM & ~LPIPS 
& ~PSNR & ~SSIM & ~LPIPS 
& ~PSNR & ~SSIM & ~LPIPS \\
\midrule
GAP-TV ~& 20.00 & 0.3680 & 0.6880 & 24.05 & 0.5660 & 0.5150 & 22.85 & 0.4060 & 0.4990 & 22.35 & 0.7660 & 0.3180 & 20.42 & 0.4260 & 0.6250 & 21.77 & 0.4320 & 0.6030 \\
PnP-FFDNet ~& 28.70 & 0.9230 & 0.1310 & 31.75 & 0.8970 & 0.1140 & 27.79 & 0.9120 & 0.1820 & 29.00 & \underline{0.9760} & 0.0510 & 29.17 & 0.9030 & 0.1190 & 28.98 & 0.8920 & 0.0980 \\
FastDVDNet ~& 29.68 & 0.9390 & 0.1040 & 32.53 & 0.9160 & 0.1050 & 28.18 & 0.9090 & 0.1750 & 29.93 & 0.9720 & 0.0520 & 29.73 & 0.9330 & 0.0980 & 30.19 & 0.9130 & 0.0790 \\
EfficientSCI ~& 33.17 & 0.9400 & 0.0450 & 32.87 & 0.9250 & 0.0700 & 30.13 & \textbf{0.9420} & 0.1120 & 30.75 & 0.9560 & 0.0460 & 32.30 & 0.9580 & 0.0600 & 31.47 & 0.9330 & 0.0480 \\
SCINeRF ~& \underline{36.40} & \underline{0.9840} & 0.0290 & 36.60 & 0.9630 & \underline{0.0220} & \underline{30.69} & 0.9330 & \underline{0.0720} & \underline{31.35} & \textbf{0.9870} & \underline{0.0310} & 33.61 & \underline{0.9630} & 0.0370 & 33.23 & \underline{0.9490} & \underline{0.0450} \\
SCIGS ~& 36.00 & 0.9640 & \underline{0.0190} & \underline{37.75} & \underline{0.9650} & 0.0290 & 27.18 & 0.7270 & 0.3000 & 29.31 & 0.9370 & 0.0810 & \underline{35.12} & 0.9580 & \underline{0.0270} & \underline{33.78} & 0.9200 & \textbf{0.0420} \\
\midrule
\rowcolor{highlightgray}[0pt][0pt] \textbf{Ours} ~& \textbf{38.52} & \textbf{0.9890} & \textbf{0.0098} & \textbf{40.04} & \textbf{0.9887} & \textbf{0.0156} & \textbf{31.36} & \underline{0.9382} & \textbf{0.0688} & \textbf{31.48} & 0.9756 & \textbf{0.0217} & \textbf{37.78} & \textbf{0.9869} & \textbf{0.0093} & \textbf{34.20} & \textbf{0.9712} & 0.0482 \\
\bottomrule
\end{tabular}%
}
\end{table*}

\begin{figure*}[!t]
    \centering
    \includegraphics[width=\textwidth]{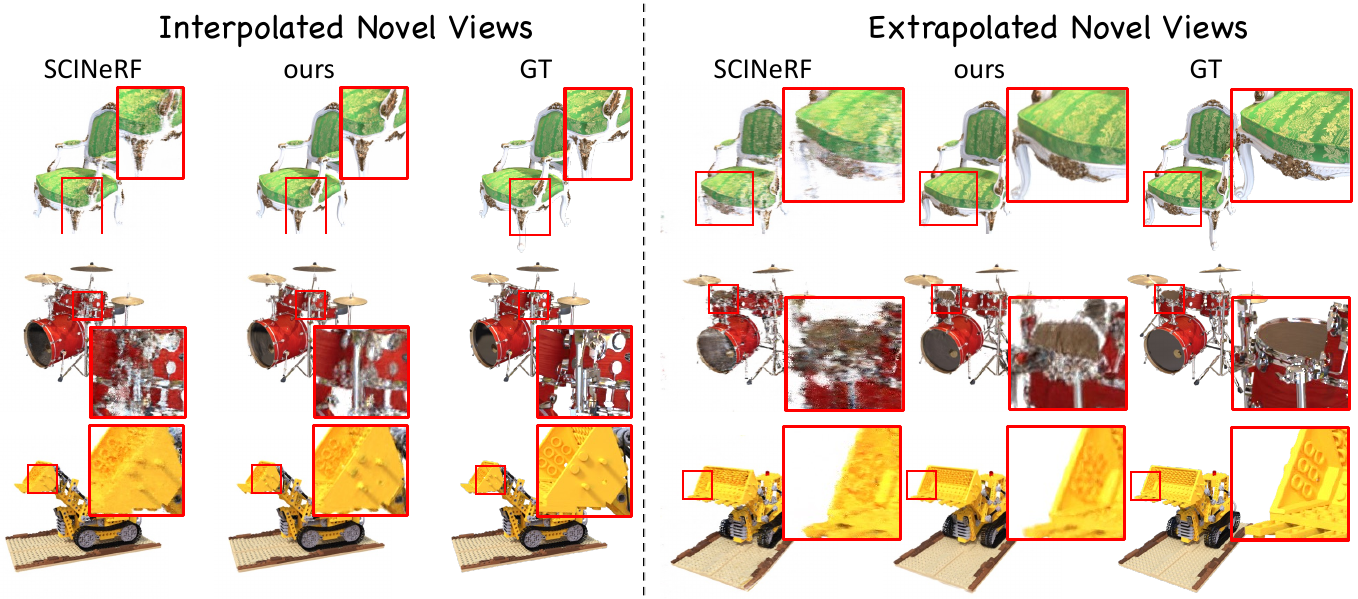}
    \caption{Comparison of novel view synthesis results with the state-of-the-art 3D-based SCI method SCINeRF~\cite{li2024scinerfneuralradiancefields}. Our method produces more accurate images under both interpolated and extrapolated viewpoints, demonstrating improved generalization to unseen views. Best viewed by zooming in.}
    \label{fig:novelview}
\end{figure*}

\paragraph{Novel View Synthesis}
We further evaluate novel-view synthesis from SCI inputs. As shown in Fig.~\ref{fig:novelview}, our method produces more coherent structures and sharper appearance than state-of-the-art SCINeRF~\cite{li2024scinerfneuralradiancefields} at unseen viewpoints. The visual comparison reflects the complete reconstruction pipeline, whereas Table~\ref{tab:ablation_fine_stage} reports a component study of the auxiliary 2D VFM refinement, with all variants initialized from the same coarse-stage reconstruction.

\subsection{Additional Results on Challenging SCI Settings}
To further evaluate the generalizability of our method beyond the standard benchmark settings, we consider additional static SCI scenes, complex measurements involving unbounded scenes and challenging camera trajectories, varying mask ratios, and high compression ratios. Dynamic-scene behavior is discussed separately in Sec.~\ref{sec:limitations}.

In addition to the state-of-the-art SCI-based 3D reconstruction method SCINeRF~\cite{li2024scinerfneuralradiancefields}, we construct three controlled 3DGS-based variants within the same SCI reconstruction framework. SCI-3DGS, SCI-MCMC, and SCI-RevADC adopt the adaptive density control and associated optimization rules of the original 3DGS~\cite{kerbl20233dgaussiansplattingrealtime}, Gaussian MCMC~\cite{kheradmand2024gaussianmcmc}, and Revising-GS~\cite{bulo2024revisingdensificationgaussiansplatting}, respectively, whereas our method uses OSGR. All four Gaussian methods otherwise share the same SCI data-fidelity objective, VFM initialization, and camera-pose refinement protocol. This construction distinguishes the contribution of OSGR from that of the shared reconstruction framework. We additionally report a separate coarse-stage novel-view synthesis comparison using the same set of methods in Table~\ref{tab:novel_view_comparison}, where our method achieves the best performance across all reported metrics.

\begin{table}[t]
\caption{Average coarse-stage novel-view synthesis comparison. Gaussian methods are evaluated before auxiliary 2D VFM refinement; SCINeRF is an external reference. Higher PSNR and SSIM and lower LPIPS indicate better performance. Bold indicates the best result in each column.}
\label{tab:novel_view_comparison}
\centering
\small
\setlength{\tabcolsep}{7pt}
\begin{tabular}{@{}lccc@{}}
\toprule
Method & PSNR$\uparrow$ & SSIM$\uparrow$ & LPIPS$\downarrow$ \\
\midrule
SCINeRF & 26.61 & 0.9360 & 0.0835 \\
SCI-3DGS & 25.23 & 0.9238 & 0.0806 \\
SCI-MCMC & 25.20 & 0.9275 & 0.0932 \\
SCI-RevADC & 25.72 & 0.9312 & 0.0696 \\
\rowcolor{highlightgray}
\textbf{Ours} & \textbf{27.71} & \textbf{0.9494} & \textbf{0.0439} \\
\bottomrule
\end{tabular}
\end{table}

All controlled Gaussian comparisons in this subsection, including the novel-view comparison above, use only the first-stage reconstructions. The second-stage pseudo-view targets are generated from and conditioned on the corresponding first-stage output; applying this refinement independently to each variant would therefore entangle the effect of its density-control strategy with differences introduced by its own pseudo-view targets. We exclude the second stage from these comparisons to preserve clear attribution.

We next consider more challenging SCI measurements that go beyond the standard bounded-scene setting. As shown in Table~\ref{tab:complex_measurements}, our method achieves the strongest overall results and ranks first or second in every entry against SCINeRF, SCI-3DGS, SCI-MCMC, and SCI-RevADC. This demonstrates robustness under more ambiguous and realistic SCI acquisition conditions and shows that the improvement is not specific to a single baseline or density-control strategy.

\begin{table*}[t]
\caption{Quantitative Comparison on Complex SCI Measurements over different 3DGS strategies.
We evaluate on more challenging SCI measurements, including unbounded scenes and measurements generated under more complex camera trajectories from Tanks and Temples~\cite{knapitsch2017tanks}, Mip-NeRF 360~\cite{barron2022mip} and DTU~\cite{jensen2014large}. To ensure a fair comparison of density-control strategies, all Gaussian variants are evaluated using first-stage reconstructions only. Our method achieves the best overall performance among these variants and the state-of-the-art baseline. \textbf{Bold} and \underline{underlined} values denote the best and second-best results, respectively.}
\label{tab:complex_measurements}
\centering
\small
\resizebox{\textwidth}{!}{%
\begin{tabular}{@{}l*{18}{@{\hspace{1pt}}c}@{}}
\toprule
Method &
\multicolumn{3}{c}{Museum} &
\multicolumn{3}{c}{Truck} &
\multicolumn{3}{c}{Bicycle} &
\multicolumn{3}{c}{Counter} &
\multicolumn{3}{c}{Garden} &
\multicolumn{3}{c}{Buda} \\
& ~PSNR & ~SSIM & ~LPIPS
& ~PSNR & ~SSIM & ~LPIPS
& ~PSNR & ~SSIM & ~LPIPS
& ~PSNR & ~SSIM & ~LPIPS
& ~PSNR & ~SSIM & ~LPIPS
& ~PSNR & ~SSIM & ~LPIPS \\
\midrule
SCINeRF & 23.92 & 0.7280 & 0.5300 & 21.11 & 0.6800 & 0.4690 & 19.67 & 0.5600 & 0.4960 & 22.01 & 0.6620 & 0.5030 & \underline{20.33} & 0.5580 & 0.5180 & \underline{28.80} & 0.8900 & 0.2620 \\
SCI-3DGS & 25.14 & \underline{0.7928} & 0.2942 & 21.68 & 0.7526 & 0.2322 & 19.40 & 0.5756 & 0.3879 & 25.18 & 0.7930 & 0.2428 & \underline{20.33} & \underline{0.6002} & \underline{0.3159} & 28.44 & \underline{0.8935} & 0.1926 \\
SCI-MCMC & \underline{27.48} & \textbf{0.8809} & \textbf{0.1431} & \underline{22.42} & \underline{0.7938} & \underline{0.1717} & \underline{20.48} & \underline{0.6237} & \underline{0.3501} & \textbf{26.28} & \textbf{0.8430} & \textbf{0.1669} & 19.63 & 0.5567 & 0.3336 & 28.04 & 0.8914 & \underline{0.1735} \\
SCI-RevADC & 25.07 & 0.7904 & 0.2970 & 21.73 & 0.7513 & 0.2318 & 19.35 & 0.5720 & 0.3840 & 25.16 & 0.7917 & 0.2420 & 20.18 & 0.5949 & 0.3167 & 28.08 & 0.8791 & 0.2241 \\
\rowcolor{highlightgray}[0pt][0pt] \textbf{Ours} & \textbf{27.52} & \textbf{0.8809} & \underline{0.1627} & \textbf{23.97} & \textbf{0.8460} & \textbf{0.1306} & \textbf{20.63} & \textbf{0.6389} & \textbf{0.3470} & \underline{26.18} & \underline{0.8377} & \underline{0.1846} & \textbf{21.60} & \textbf{0.6622} & \textbf{0.2816} & \textbf{30.15} & \textbf{0.9231} & \textbf{0.1272} \\
\bottomrule
\end{tabular}%
}
\end{table*}

\begin{figure*}[!t]
    \centering
    \includegraphics[width=\textwidth]{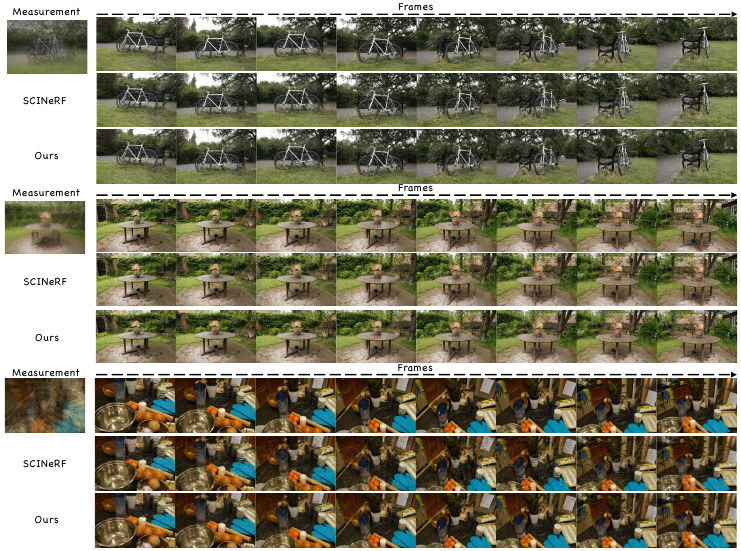}
    \caption{Qualitative results under complex viewpoint changes. The top and middle examples cover wide-range viewpoint changes, while the bottom example contains irregular sweeping motion and large transformations. From top to bottom: ground truth, SCINeRF~\cite{li2024scinerfneuralradiancefields}, and our method.}
    \label{fig:complexview}
\end{figure*}

\paragraph{Complex Viewpoint Variations}
We further evaluate challenging SCI scenarios involving irregular and large-scale camera motion from Mip-NeRF 360~\cite{barron2022mip}. As shown in Fig.~\ref{fig:complexview}, SCINeRF~\cite{li2024scinerfneuralradiancefields} struggles to preserve structural coherence under these viewpoint changes, leading to distortions and multi-view inconsistency. In contrast, our method produces more accurate and consistent reconstructions by exploiting the complementary geometric and image priors of the VFMs. Together with the quantitative results in Table~\ref{tab:complex_measurements}, this comparison shows that our method achieves the best overall performance against SCINeRF, SCI-3DGS, SCI-MCMC, and SCI-RevADC under complex viewpoint variations.

\paragraph{Robustness to Mask Ratio}
We next vary the mask ratio while retaining the same coarse-stage evaluation protocol. Table~\ref{tab:mask_ratio} compares SCINeRF, SCI-3DGS, SCI-MCMC, SCI-RevADC, and our method at every tested ratio from 0.125 to 0.75. The controlled Gaussian variants allow us to distinguish robustness arising from the proposed density-control design from that of the shared 3DGS reconstruction framework. Our method achieves the best performance at every mask ratio, outperforming SCINeRF and all three controlled variants, including under the more challenging extreme settings. These results demonstrate that the proposed reconstruction remains robust to changes in both the spatial coding pattern and the amount of information retained by the SCI measurement throughout the evaluated mask-ratio range.

\begin{table}[!t]
\caption{Coarse-stage base-view reconstruction under different mask ratios with $N=8$ encoded views.}
\label{tab:mask_ratio}
\centering
\small
\setlength{\tabcolsep}{4pt}
\renewcommand{\arraystretch}{1.05}
\begin{tabular}{
c@{\hspace{8pt}}
l@{\hspace{4pt}}
c@{\hspace{5pt}}c@{\hspace{5pt}}c
}
\toprule
Mask Ratio & Method & PSNR$\uparrow$ & SSIM$\uparrow$ & LPIPS$\downarrow$ \\
\midrule
\multirow{5}{*}{$0.125$} & SCINeRF & 32.31 & 0.9595 & 0.0675 \\
& SCI-3DGS & 31.69 & 0.9541 & 0.0632 \\
& SCI-MCMC & 32.92 & 0.9638 & 0.0495 \\
& SCI-RevADC & 31.77 & 0.9551 & 0.0622 \\
\rowcolor{highlightgray} & \textbf{Ours} & \textbf{34.10} & \textbf{0.9713} & \textbf{0.0334} \\
\cmidrule(lr){1-5}
\multirow{5}{*}{$0.25$} & SCINeRF & 33.65 & 0.9632 & 0.0393 \\
& SCI-3DGS & 33.33 & 0.9580 & 0.0606 \\
& SCI-MCMC & 34.43 & 0.9640 & 0.0477 \\
& SCI-RevADC & 33.36 & 0.9572 & 0.0625 \\
\rowcolor{highlightgray} & \textbf{Ours} & \textbf{35.57} & \textbf{0.9751} & \textbf{0.0289} \\
\cmidrule(lr){1-5}
\multirow{5}{*}{$0.5$} & SCINeRF & 34.38 & 0.9648 & 0.0496 \\
& SCI-3DGS & 32.85 & 0.9500 & 0.0721 \\
& SCI-MCMC & 33.44 & 0.9579 & 0.0588 \\
& SCI-RevADC & 33.35 & 0.9548 & 0.0670 \\
\rowcolor{highlightgray} & \textbf{Ours} & \textbf{34.47} & \textbf{0.9668} & \textbf{0.0468} \\
\cmidrule(lr){1-5}
\multirow{5}{*}{$0.75$} & SCINeRF & 32.94 & 0.9315 & 0.0976 \\
& SCI-3DGS & 32.23 & 0.9388 & 0.0863 \\
& SCI-MCMC & 32.81 & 0.9477 & 0.0731 \\
& SCI-RevADC & 33.03 & 0.9492 & 0.0681 \\
\rowcolor{highlightgray} & \textbf{Ours} & \textbf{34.24} & \textbf{0.9618} & \textbf{0.0490} \\
\bottomrule
\end{tabular}
\end{table}

\paragraph{High Compression Ratios}
We then evaluate high-compression SCI measurements using scenes from DTU~\cite{jensen2014large} and Mip-NeRF 360~\cite{barron2022mip}, with the mask ratio fixed at 0.25. Table~\ref{tab:high_compression} includes SCINeRF together with SCI-3DGS, SCI-MCMC, and SCI-RevADC, enabling a consistent comparison of representation and density-control choices at $\mathrm{CR}=16$ and $\mathrm{CR}=32$. Our method achieves the best overall performance at both compression ratios, outperforming SCINeRF and all three controlled variants. The advantage remains pronounced at $\mathrm{CR}=32$, showing that the proposed priors and SCI-specific optimization remain effective when substantially more views are multiplexed into a single measurement.

\begin{table}[!t]
\caption{Coarse-stage base-view reconstruction under high compression ratios on scenes from DTU~\cite{jensen2014large} and Mip-NeRF 360~\cite{barron2022mip}. The mask ratio is fixed at 0.25.}
\label{tab:high_compression}
\centering
\small
\setlength{\tabcolsep}{4pt}
\renewcommand{\arraystretch}{1.05}
\begin{tabular}{
c@{\hspace{8pt}}
l@{\hspace{4pt}}
c@{\hspace{5pt}}c@{\hspace{5pt}}c
}
\toprule
Compression & Method & PSNR$\uparrow$ & SSIM$\uparrow$ & LPIPS$\downarrow$ \\
\midrule
\multirow{5}{*}{$\mathrm{CR}=16$} & SCINeRF & 21.60 & 0.5697 & 0.5281 \\
& SCI-3DGS & 22.33 & 0.6295 & 0.4292 \\
& SCI-MCMC & 23.15 & 0.6541 & 0.3709 \\
& SCI-RevADC & 22.26 & 0.6274 & 0.4229 \\
\rowcolor{highlightgray} & \textbf{Ours} & \textbf{23.65} & \textbf{0.6844} & \textbf{0.3688} \\
\cmidrule(lr){1-5}
\multirow{5}{*}{$\mathrm{CR}=32$} & SCINeRF & 18.69 & 0.4647 & 0.6033 \\
& SCI-3DGS & 21.02 & 0.5599 & 0.5032 \\
& SCI-MCMC & 21.28 & 0.5708 & 0.4821 \\
& SCI-RevADC & 20.94 & 0.5563 & 0.5034 \\
\rowcolor{highlightgray} & \textbf{Ours} & \textbf{21.85} & \textbf{0.5964} & \textbf{0.4803} \\
\bottomrule
\end{tabular}
\end{table}

\paragraph{Camera Trajectory Accuracy}
Accurate camera poses are essential for geometrically consistent 3D reconstruction from SCI. We compare trajectories estimated by SCINeRF, SCI-3DGS, SCI-MCMC, SCI-RevADC, and ours on scenes from NeRF Synthetic~\cite{mildenhall2020nerfrepresentingscenesneural} and Mip-NeRF 360~\cite{barron2022mip}. Table~\ref{tab:trajectory_accuracy} reports ATE, $\mathrm{RPE}_t$, $\mathrm{RPE}_r$, $\mathrm{nATE}$, and $\mathrm{nRPE}_t$ as complementary global and local trajectory errors. Ours achieves the lowest value for every metric among SCINeRF and the three controlled Gaussian variants.

\begin{table*}[!t]
\caption{Average camera trajectory accuracy on scenes from NeRF Synthetic~\cite{mildenhall2020nerfrepresentingscenesneural} and Mip-NeRF 360~\cite{barron2022mip}. $\mathrm{nATE}$ and $\mathrm{nRPE}_t$ denote normalized ATE and normalized $\mathrm{RPE}_t$, respectively. Lower is better for all metrics.}
\label{tab:trajectory_accuracy}
\centering
\small
\setlength{\tabcolsep}{6pt}
\renewcommand{\arraystretch}{1.08}
\begin{tabular}{@{}lccccccc@{}}
\toprule
Method & ATE$\downarrow$ & $\mathrm{nATE}\downarrow$ & Trans. mean$\downarrow$ & Rot. mean ($^\circ$)$\downarrow$ & $\mathrm{RPE}_t\downarrow$ & $\mathrm{nRPE}_t\downarrow$ & $\mathrm{RPE}_r$ ($^\circ$)$\downarrow$ \\
\midrule
SCINeRF & 0.149424 & 0.050720 & 0.135576 & 45.7417 & 0.394496 & 0.092206 & 4.7359 \\
SCI-3DGS & 0.044590 & 0.014153 & 0.040852 & 4.1965 & 0.048161 & 0.015108 & 1.0580 \\
SCI-MCMC & 0.068249 & 0.021450 & 0.062590 & 6.0624 & 0.081741 & 0.025888 & 2.0724 \\
SCI-RevADC & 0.042689 & 0.013698 & 0.038905 & 3.3509 & 0.040692 & 0.013089 & 0.9033 \\
\rowcolor{highlightgray}
\textbf{Ours} & \textbf{0.042353} & \textbf{0.013490} & \textbf{0.038785} & \textbf{3.3018} & \textbf{0.039854} & \textbf{0.012780} & \textbf{0.8927} \\
\bottomrule
\end{tabular}
\end{table*}

\subsection{Ablation Study}

\begin{figure*}[!t]
    \centering
    \includegraphics[width=\textwidth]{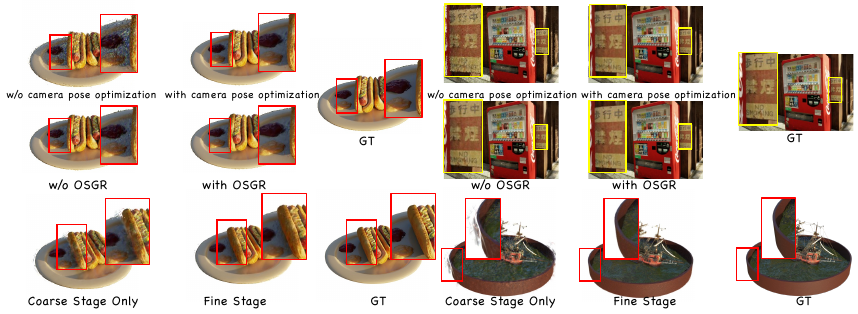}
    \caption{Qualitative ablations of camera-pose optimization, OSGR, and auxiliary 2D VFM refinement. Removing camera-pose optimization or OSGR produces broader changes in geometry and sharpness, whereas the differences associated with auxiliary refinement are concentrated in local structure and appearance. The complete configuration combines these complementary effects.}
    \label{fig:ablation_2d}
\end{figure*}

\begin{table*}[!t]
\fontsize{9.5}{10}\selectfont
\setlength{\tabcolsep}{5pt}
\renewcommand{\arraystretch}{0.9}
\caption{Component ablation on six scenes. A \cmark\ denotes an enabled component. OOM denotes incomplete uncapped runs.}
\label{tab:ablation-vfm-check}
\centering
\begin{tabular}{@{}ccccccccc@{}}
\toprule
\shortstack{VFM\\Points} & \shortstack{VFM\\Poses} & \shortstack{Opacity\\Split} & \shortstack{Opacity\\Reg.} & \shortstack{Count\\Cap} & \shortstack{Pose\\Refine} & PSNR$\uparrow$ & SSIM$\uparrow$ & LPIPS$\downarrow$ \\
\midrule
\xmark & \xmark & \xmark & \xmark & \xmark & \xmark & 21.20 & 0.6980 & 0.4490 \\
\xmark & \cmark & \cmark & \cmark & \cmark & \cmark & 33.51 & 0.9652 & 0.0436 \\
\cmark & \xmark & \cmark & \cmark & \cmark & \cmark & 30.30 & 0.9404 & 0.0777 \\
\cmark & \cmark & \xmark & \cmark & \cmark & \cmark & 33.26 & 0.9570 & 0.0629 \\
\cmark & \cmark & \cmark & \xmark & \cmark & \cmark & 33.90 & 0.9619 & 0.0604 \\
\cmark & \cmark & \cmark & \cmark & \xmark & \cmark & \multicolumn{3}{c}{OOM} \\
\cmark & \cmark & \cmark & \cmark & \cmark & \xmark & 29.70 & 0.9090 & 0.1476 \\
\rowcolor{highlightgray} \cmark & \cmark & \cmark & \cmark & \cmark & \cmark & \textbf{35.57} & \textbf{0.9751} & \textbf{0.0289} \\
\bottomrule
\end{tabular}
\end{table*}
Table~\ref{tab:ablation-vfm-check} separately evaluates opacity-guided splitting, mean-opacity regulation, and the Gaussian-count constraint. Without the count constraint, a subset of runs exhausts GPU memory before 20,000 iterations, leaving no complete six-scene aggregate. Each remaining component improves all three reconstruction metrics, and Fig.~\ref{fig:ablation_2d} illustrates their combined qualitative effect.

\paragraph{Comparison of 3DGS Densification Strategies}

Adaptive density control rules developed for dense multi-view supervision may behave differently when multiple views are entangled in one SCI measurement. We therefore report two complementary comparisons in Tables~\ref{tab:without_novel_compare} and~\ref{tab:scisplat_reproduction}.

\emph{Controlled full-20 comparison.} Table~\ref{tab:without_novel_compare} includes all 20 cases and is designed to isolate adaptive density control and method-specific optimization. SCI-3DGS, SCI-MCMC, SCI-RevADC, and ours use the same SCI data-fidelity objective, VFM initialization, camera-pose refinement, training budget, and evaluation data; only their adaptive density control and associated optimization rules differ, following vanilla 3DGS~\cite{kerbl20233dgaussiansplattingrealtime}, MCMC~\cite{kheradmand2024gaussianmcmc}, Revising-GS~\cite{bulo2024revisingdensificationgaussiansplatting}, and OSGR, respectively. All Gaussian variants are evaluated after the first stage. We omit the fine stage because its pseudo-view targets are generated from the corresponding first-stage reconstruction, so applying it independently could propagate coarse-stage differences and obscure attribution. SCINeRF is included only as an external reference. Thus, SCI-MCMC measures the effect of the original MCMC rules inside the shared protocol. Under this controlled setting, ours achieves the best reconstruction quality, renders at 407.944 FPS, and completes training in 53.67 minutes. SCI-RevADC is marginally faster at 415.343 FPS and 50.24 minutes, but has substantially worse PSNR, SSIM, and LPIPS.

\emph{Matched 17-case comparison at the coarse stage.} Because SCISplat$^{\dagger}$~\cite{li2024learningradiancefieldssingle} has no public implementation, we reproduce its published pipeline under the same data and hardware protocol. We attempt the same 20 cases, but SCISplat$^{\dagger}$ fails to initialize on three. Table~\ref{tab:scisplat_reproduction} therefore reports paired averages over the 17 successful cases, with ours recomputed on exactly the same subset before auxiliary 2D VFM refinement. Because each refinement target is generated from its corresponding first-stage output, applying the second stage separately can propagate and compound existing coarse-stage differences. We therefore compare first-stage results only, preserving direct attribution to the reconstruction method. Under this protocol, ours improves PSNR by 3.29~dB and SSIM by 0.0538 while reducing LPIPS by 0.0694 and training time by 32.16 minutes, and it yields lower error on every trajectory metric.

\begin{table*}[!t]
\small
\caption{Controlled first-stage comparison over all 20 cases. The four Gaussian variants are recomputed with the same SCI objective, VFM initialization, pose-refinement protocol, and training budget; SCINeRF is an external reference. FPS and training time use the fixed-hardware benchmark. Bold values indicate the best result in each column.}
\label{tab:without_novel_compare}
\centering
\setlength{\tabcolsep}{7pt}
\begin{tabular}{lccccc}
\toprule
Method & PSNR$\uparrow$ & SSIM$\uparrow$ & LPIPS$\downarrow$ & FPS$\uparrow$ & Training Time (min)$\downarrow$ \\
\midrule
SCINeRF & 27.84 & 0.8483 & 0.2215 & 0.372 & 746.84 \\
SCI-3DGS & 27.80 & 0.8619 & 0.1594 & 382.263 & 54.33 \\
SCI-MCMC & 29.05 & 0.8821 & 0.1276 & 284.647 & 59.20 \\
SCI-RevADC & 27.92 & 0.8631 & 0.1566 & \textbf{415.343} & \textbf{50.24} \\
\rowcolor{highlightgray}
Ours & \textbf{29.80} & \textbf{0.9005} & \textbf{0.1090} & 407.944 & 53.67 \\
\bottomrule
\end{tabular}
\end{table*}

\begin{table*}[!t]
\caption{Coarse-stage comparison with SCISplat$^{\dagger}$~\cite{li2024learningradiancefieldssingle}, where $^{\dagger}$ denotes our reproduction. Reconstruction and training results average 17 successful cases (three of 20 fail). Bold indicates the better result.}
\label{tab:scisplat_reproduction}
\centering
\small
\setlength{\tabcolsep}{4pt}
\begin{tabular}{@{}l*{9}{c}@{}}
\toprule
& \multicolumn{4}{c}{Reconstruction and Training}
& \multicolumn{5}{c}{Trajectory Accuracy} \\
\cmidrule(lr){2-5}\cmidrule(lr){6-10}
Method & PSNR$\uparrow$ & SSIM$\uparrow$ & LPIPS$\downarrow$ & Train (min)$\downarrow$
& ATE RMSE$\downarrow$ & $\mathrm{nATE}\downarrow$ & Rotation ($^\circ$)$\downarrow$
& $\mathrm{nRPE}_t\downarrow$ & $\mathrm{RPE}_r$ ($^\circ$)$\downarrow$ \\
\midrule
SCISplat$^{\dagger}$ & 26.79 & 0.8461 & 0.1761 & 86.89
& 0.08023 & 0.04062 & 16.767 & 0.04446 & 1.534 \\
\rowcolor{highlightgray}
\textbf{Ours} & \textbf{30.08} & \textbf{0.8999} & \textbf{0.1067} & \textbf{54.73}
& \textbf{0.04235} & \textbf{0.01349} & \textbf{3.302} & \textbf{0.01278} & \textbf{0.893} \\
\bottomrule
\end{tabular}
\end{table*}

\begin{table}[!t]
\caption{Ablation of auxiliary 2D VFM refinement and alpha-based support weighting over eight scenes. All variants share the same coarse reconstruction and are evaluated at held-out poses disjoint from those used for pseudo-view supervision.}
\label{tab:ablation_fine_stage}
\centering
\small
\setlength{\tabcolsep}{5pt}
\begin{tabular}{@{}llccc@{}}
\toprule
View Set & Method & PSNR$\uparrow$ & SSIM$\uparrow$ & LPIPS$\downarrow$ \\
\midrule
\multirow{3}{*}{All Novel} & Coarse Output & 27.71 & 0.9494 & 0.0439 \\
& 2D VFM & 27.75 & 0.9498 & 0.0426 \\
& \cellcolor{highlightgray}\textbf{+ Support Weight} & \cellcolor{highlightgray}\textbf{27.76} & \cellcolor{highlightgray}\textbf{0.9498} & \cellcolor{highlightgray}\textbf{0.0425} \\
\midrule
\multirow{3}{*}{Extrapolated} & Coarse Output & 25.19 & 0.9273 & 0.0779 \\
& 2D VFM & 25.38 & 0.9295 & 0.0720 \\
& \cellcolor{highlightgray}\textbf{+ Support Weight} & \cellcolor{highlightgray}\textbf{25.39} & \cellcolor{highlightgray}\textbf{0.9297} & \cellcolor{highlightgray}\textbf{0.0715} \\
\bottomrule
\end{tabular}
\end{table}

\paragraph{Auxiliary 2D VFM Refinement}
We evaluate the auxiliary refinement from the same coarse reconstruction. As shown in Table~\ref{tab:ablation_fine_stage}, unweighted 2D VFM supervision improves all aggregate metrics, particularly on extrapolated views. Alpha-based support weighting further maintains or improves every metric and yields the highest PSNR and lowest LPIPS for both view sets. The 2D VFM thus supplies refinement targets, while the support weight modulates their spatial influence according to the accumulated alpha of the current Gaussian representation. Figure~\ref{fig:ablation_2d} illustrates the corresponding changes in local structure and appearance.

\paragraph{Qualitative Analysis of OSGR}
We further provide both 2D and 3D visual ablations to analyze the effect of the proposed components. In the 2D rendering comparison shown in Fig.~\ref{fig:ablation_2d}, removing OSGR or camera pose optimization leads to noticeably blurrier reconstructions and local structural distortions, whereas the full model recovers sharper details and more faithful appearance. These visualizations further support that OSGR improves both image-space fidelity and 3D structural consistency under the compressed SCI setting.

\paragraph{Effect of Opacity Regulation}
We evaluate the mean-opacity term with the remaining OSGR components fixed, including the Gaussian-count constraint. As summarized in Table~\ref{tab:ablation-vfm-check} and further illustrated by the population trajectories in the supplementary material, removing this term yields lower reconstruction quality and a substantially larger Gaussian population by the 7,000-iteration reference point. The mean-opacity term provides soft regulation during active densification. After the reference iteration, the count constraint fixes the population reached at that point as the hard upper bound for subsequent splitting and duplication. Together, they enable the complete model to maintain a substantially more compact representation than the variant without opacity regulation while achieving the best reconstruction metrics.

\subsection{Limitations and Future Work}
\label{sec:limitations}

\begin{table}[!b]
 \caption{Quantitative comparison on the Bear and Flamingo dynamic scenes from DAVIS~\cite{ponttuset20182017davischallengevideo}. All Gaussian variants are evaluated using their complete fine-stage outputs.}
\label{tab:dynamic_sci}
\centering
\small
\setlength{\tabcolsep}{4pt}
\renewcommand{\arraystretch}{0.95}
\begin{tabular}{
c@{\hspace{8pt}}
l@{\hspace{4pt}}
c@{\hspace{5pt}}c@{\hspace{5pt}}c
}
\toprule
Scene & Method & PSNR$\uparrow$ & SSIM$\uparrow$ & LPIPS$\downarrow$ \\
\midrule
\multirow{5}{*}{Bear} & SCINeRF & 23.96 & 0.8560 & 0.1940 \\
& SCI-3DGS & 24.10 & 0.8119 & 0.1671 \\
& SCI-MCMC & \textbf{27.46} & \textbf{0.9072} & \textbf{0.0843} \\
& SCI-RevADC & 25.16 & 0.8499 & 0.1345 \\
\rowcolor{highlightgray} & \textbf{Ours} & 26.90 & 0.8923 & 0.0997 \\
\cmidrule(lr){1-5}
\multirow{5}{*}{Flamingo} & SCINeRF & 25.11 & 0.8240 & 0.2700 \\
& SCI-3DGS & 25.13 & 0.8313 & 0.2237 \\
& SCI-MCMC & \textbf{28.90} & \textbf{0.9115} & \textbf{0.1112} \\
& SCI-RevADC & 25.20 & 0.8306 & 0.2150 \\
\rowcolor{highlightgray} & \textbf{Ours} & 27.59 & 0.8949 & 0.1396 \\
\bottomrule
\end{tabular}
\end{table}

Table~\ref{tab:dynamic_sci} evaluates two DAVIS sequences outside our static-scene formulation~\cite{ponttuset20182017davischallengevideo}. For SCI-3DGS, SCI-MCMC, SCI-RevADC, and ours, we report the corresponding complete fine-stage outputs. Ours obtains better values than \mbox{SCINeRF}~\cite{li2024scinerfneuralradiancefields}, SCI-3DGS, and SCI-RevADC in all six scene--metric entries, but SCI-MCMC achieves the best value for every reported metric. Ours ranks second throughout, indicating that dynamic scenes remain a limitation of the current formulation.

One plausible OSGR-specific hypothesis concerns the use of local opacity contrast as an additional splitting cue. Independent object motion makes the static forward model misspecified: one Gaussian set cannot consistently explain all masked views multiplexed into the measurement. Under this mismatch, optimized opacity may reflect not only persistent surface support but also compensation for motion-induced residuals. A compensatory primitive supported by only a subset of views may become a local opacity outlier and satisfy OSGR's splitting criterion despite lacking temporal support. The transmittance-preserving child update means that splitting does not by itself imply an immediate increase in accumulated opacity, while the global opacity penalty and Gaussian-count constraint limit proliferation. However, none of these mechanisms identifies whether a local opacity peak arises from static geometry or motion compensation. Repeated selection could therefore divert bounded representation capacity toward ambiguous dynamic regions and potentially preserve ghost contours or fragmented geometry after subsequent optimization. Future work will incorporate motion-aware cross-view support into splitting and extend the representation to 4D Gaussians~\cite{wu20244dgaussiansplattingrealtime}, thereby improving robustness in general dynamic SCI settings under complex object motion.

\section{Conclusion}

We present a framework for 3D scene reconstruction from a single SCI measurement by combining vision foundation models with a tailored 3DGS-based pipeline. Our primary reconstruction combines measurement-derived 3D VFM initialization with SCI-aware Gaussian optimization, where OSGR combines opacity-guided splitting with global opacity regulation to stabilize Gaussian growth under sparse SCI supervision. An auxiliary 2D VFM stage provides complementary pseudo-view supervision at synthesized viewpoints for local appearance refinement. Extensive experiments across diverse SCI settings demonstrate leading performance in both reconstruction and novel-view synthesis. More broadly, the modular design permits increasingly capable geometry and image foundation models to be incorporated into the corresponding initialization and refinement stages, providing a practical path toward higher-quality 3D reconstruction from compressed visual observations and more challenging SCI acquisition settings.

\FloatBarrier
\bibliographystyle{IEEEtran}
\bibliography{main}

\clearpage
\appendices
\makeatletter
\let\maketitle\GSorigmaketitle
\let\@maketitle\GSorigatmaketitle
\let\thanks\GSorigthanks
\makeatother
\title{GS$^{\mathbf{2}}$CI: Robust Gaussian Splatting For \\
Snapshot Compressive Imaging via Large Vision Model Priors \\
Supplementary Material}
\author{Yanming~Yang, Chenxi~Song, Ping~Wang, Xin~Yuan, and Chi~Zhang%
\thanks{\hspace*{-1em}Corresponding author: Chi Zhang (chizhang@westlake.edu.cn). \\ Yanming Yang, Chenxi Song and Chi Zhang are with the AGI Lab, Westlake University, Hangzhou 310030, China. Ping Wang and Xin Yuan are with Westlake University, Hangzhou 310030, China. (e-mail: \{yangyanming, songchenxi, wangping, xylab, chizhang\}@westlake.edu.cn).}}
\maketitle
\newcommand{\MainFigref}[2]{Fig.~#2 in the main paper}
\newcommand{\MainTabref}[2]{Table~#2 in the main paper}
\newcommand{\MainSecref}[2]{Sec.~#2 in the main paper}
\newcommand{\MainEqref}[2]{Eq.~(#2) in the main paper}
\providecommand{\MainFigref}[2]{Fig.~\ref{#1}}
\providecommand{\MainTabref}[2]{Table~\ref{#1}}
\providecommand{\MainSecref}[2]{Sec.~\ref{#1}}
\providecommand{\MainEqref}[2]{Eq.~\eqref{#1}}

\section{Multiplexing-Aware Geometry Initialization}
\label{sec:supp-geometry-initialization}
\enlargethispage{1pt}
VGGT~\cite{wang2025vggtvisualgeometrygrounded} assumes a sequence of view-resolved RGB images, whereas SCI collapses $N$ coded views into the single multiplexed observation $\mathbf{Y}$ defined in \MainEqref{eq:sci}{1}. Consequently, the raw SCI measurement does not satisfy VGGT's input model: structures from different views are superimposed rather than individually resolved. We address this mismatch by constructing geometry-oriented proxy views from $\mathbf{Y}$ and the known binary coding masks $\mathcal{M}=\{\mathbf{M}_i\}_{i=1}^{N}$. The construction adapts to the realized mask overlap and completes each proxy by interpolating the locations omitted by its mask before the proxy is supplied to VGGT.

Specifically, let $\mathbf{X}_i\in\mathbb{R}^{H\times W\times3}$ denote the latent RGB image of view $i$, $\mathbf{M}_i\in\{0,1\}^{H\times W}$ its corresponding coding mask shared across RGB channels, and $\mathbf{p}$ a pixel location. Our construction produces the sequence $\widetilde{\mathcal{X}}=\{\widetilde{\mathbf{X}}_i\}_{i=1}^{N}$ of RGB proxy views on the same $H\times W$ sampling grid. These proxies are supplied exclusively to the frozen VGGT model for geometry initialization; they are not interpreted as SCI reconstructions and never serve as photometric supervision.

The mean measurement multiplicity determines whether the proxy generator uses Energy-Normalized Initialization (ENI) or Adaptive Mask-Decoding Initialization (AMDI):
\begin{equation}
    \mu_{\mathcal{M}}
    =
    \frac{1}{HW}
    \sum_{\mathbf{p}}
    \sum_{i=1}^{N}\mathbf{M}_i(\mathbf{p}),
    \label{eq:supp-mask-multiplicity}
\end{equation}
which measures the average number of latent views contributing to one measurement pixel. With $\tau_{\mu}=4$ fixed for all scenes, the proxy construction is selected as follows:
\begin{equation}
    \widetilde{\mathcal{X}}
    =\mathcal{P}(\mathbf{Y},\mathcal{M})
    =
    \begin{cases}
        \mathcal{P}_{\mathrm{ENI}}(\mathbf{Y},\mathcal{M}),
        & \mu_{\mathcal{M}}<\tau_{\mu},\\
        \mathcal{P}_{\mathrm{AMDI}}(\mathbf{Y},\mathcal{M}),
        & \mu_{\mathcal{M}}\geq\tau_{\mu}.
    \end{cases}
    \label{eq:supp-proxy-routing}
\end{equation}

The two constructions address complementary mixing regimes. At low multiplicity, ENI exposes mask-supported samples without fitting a local-constant multi-view inverse model. As multiplicity increases, however, exposure normalization retains only aggregate intensity and cannot separate the superposed latent-view contributions. Conversely, AMDI's full-rank local system may require enlarged spatial support when mixing is weak. Equation~\eqref{eq:supp-proxy-routing} therefore balances cross-view ambiguity against local inverse-model bias rather than identifying a scene category.

For low-multiplicity acquisitions, ENI compensates for first-order exposure variation induced by mask overlap as follows:
\begin{equation}
\begin{gathered}
    a(\mathbf{p})=\sum_{i=1}^{N}\mathbf{M}_i(\mathbf{p}),\\
    \overline{\mathbf{Y}}(\mathbf{p})=
    \mathbf{Y}(\mathbf{p})/a(\mathbf{p}).
\end{gathered}
\label{eq:supp-energy-normalization}
\end{equation}
The division is applied wherever $a(\mathbf{p})>0$, and we set $\overline{\mathbf{Y}}(\mathbf{p})=0$ when $a(\mathbf{p})=0$.
Each view-specific proxy is then obtained by remodulating the normalized measurement with $\mathbf{M}_i$ and completing the mask-omitted locations:
\begin{equation}
    \widetilde{\mathbf{X}}^{\mathrm{ENI}}_i
    =
    \mathcal{I}\!\left(
        \overline{\mathbf{Y}}\odot\mathbf{M}_i;\mathbf{M}_i
    \right).
    \label{eq:supp-eni-proxy}
\end{equation}
For a binary mask, $\mathcal{I}$ treats the values at $\mathcal{S}_i=\{\mathbf{p}:\mathbf{M}_i(\mathbf{p})=1\}$ as known samples, linearly interpolates missing pixels inside $\operatorname{conv}(\mathcal{S}_i)$, and uses nearest-neighbor extrapolation outside $\operatorname{conv}(\mathcal{S}_i)$. Thus, each proxy retains its mask-supported samples and is completed on the original grid.

When $\mu_{\mathcal{M}}$ is high, more latent views overlap at each measurement pixel. Energy normalization corrects the aggregate exposure but does not separate these view contributions, so direct mask remodulation remains ambiguous. AMDI therefore exploits the spatial variation of the actual coding patterns to estimate local per-view contributions. Define $\mathbf{m}(\mathbf{q})=[\mathbf{M}_1(\mathbf{q}),\ldots,\mathbf{M}_N(\mathbf{q})]^{\mathsf{T}}\in\mathbb{R}^{N}$. Within a clipped square neighborhood $\Omega_s(\mathbf{p})$, AMDI adopts the local-constant model $\mathbf{Y}(\mathbf{q})\approx\mathbf{U}(\mathbf{p})^{\mathsf{T}}\mathbf{m}(\mathbf{q})+\mathbf{Z}(\mathbf{q})$, where $\mathbf{U}(\mathbf{p})\in\mathbb{R}^{N\times3}$ stacks the unknown RGB values of the $N$ proxies. Its normal-equation terms are:
\begin{equation}
\begin{aligned}
    \mathbf{G}_{s}(\mathbf{p})
    &=
    \sum_{\mathbf{q}\in\Omega_s(\mathbf{p})}
    \mathbf{m}(\mathbf{q})\mathbf{m}(\mathbf{q})^{\mathsf{T}},\\
    \mathbf{B}_{s}(\mathbf{p})
    &=
    \sum_{\mathbf{q}\in\Omega_s(\mathbf{p})}
    \mathbf{m}(\mathbf{q})\mathbf{Y}(\mathbf{q})^{\mathsf{T}}.
\end{aligned}
\label{eq:supp-local-normal}
\end{equation}
Let $\mathbf{G}_{\mathrm{all}}=\sum_{\mathbf{q}}\mathbf{m}(\mathbf{q})\mathbf{m}(\mathbf{q})^{\mathsf{T}}$ be the full-image mask Gram matrix. We select the smallest numerically full-rank and well-conditioned neighborhood:
\begin{equation}
\begin{split}
    s^{\star}(\mathbf{p})
    =\min_{s\in\mathcal{S}}\big\{s:\;&
    \operatorname{rank}_{\mathrm{num}}\!\left(\mathbf{G}_{s}(\mathbf{p})\right)=N,\\
    &\lambda_{\min}\!\left(\mathbf{G}_{s}(\mathbf{p})\right)\geq\tau_{\lambda},\\
    &\kappa_{2}\!\left(\mathbf{G}_{s}(\mathbf{p})\right)
    \leq\gamma\,\kappa_{2}(\mathbf{G}_{\mathrm{all}})\big\},
\end{split}
\label{eq:supp-adaptive-window}
\end{equation}
where $\kappa_2$ is the spectral condition number, $\tau_{\lambda}=1$, and $\gamma=2$. Numerical rank is evaluated at the solver tolerance. Given the selected unregularized Gram matrix, a trace-normalized ridge produces the proxy stack:
\begin{equation}
\begin{aligned}
    \lambda^{\mathrm{ridge}}_{\mathbf{p}}
    &=
    \eta\,
    \frac{\operatorname{tr}\!\left(\mathbf{G}_{s^{\star}}(\mathbf{p})\right)}{N},\\
    \widetilde{\mathbf{X}}^{\mathrm{AMDI}}(\mathbf{p})
    &=
    \left(
        \mathbf{G}_{s^{\star}}(\mathbf{p})
        +\lambda^{\mathrm{ridge}}_{\mathbf{p}}\mathbf{I}_{N}
    \right)^{-1}
    \mathbf{B}_{s^{\star}}(\mathbf{p}).
\end{aligned}
\label{eq:supp-amdi-decode}
\end{equation}
We set $\eta=10^{-3}$. At each pixel, the $N$ rows of $\widetilde{\mathbf{X}}^{\mathrm{AMDI}}(\mathbf{p})$ provide the RGB values of the $N$ proxy views. The evaluated neighborhood sizes are:
\begin{equation}
\begin{split}
    \mathcal{S}=\{&3,5,7,9,11,13,15,17,\\
                  &21,25,31,41,57,81\}.
\end{split}
\label{eq:supp-window-schedule}
\end{equation}
Boundary windows contain only unique in-image samples, without padding or repetition. The listed sizes cover all experiments; for larger inputs, odd window sizes grow geometrically until full-image coverage. AMDI requires an admissible full-image system and a neighborhood satisfying Eq.~\eqref{eq:supp-adaptive-window}.

For the acquisition settings evaluated in \MainTabref{tab:mask_ratio}{IV} and \MainTabref{tab:high_compression}{V}, Eq.~\eqref{eq:supp-proxy-routing} assigns the 8-view mask-ratio settings of 0.125 and 0.25 to ENI, those of 0.5 and 0.75 to AMDI, and the $\mathrm{CR}=16$ and $\mathrm{CR}=32$ settings at mask ratio 0.25 to AMDI. Both branches produce $N$ complete RGB proxy views before VGGT preprocessing. The local neighborhoods introduced by AMDI are used only to construct these views; VGGT is always applied to the complete proxy sequence. The choice between ENI and AMDI is determined solely by the acquisition masks through Eq.~\eqref{eq:supp-proxy-routing}, without reference imagery, scene-dependent information, or external camera calibration.

The proxy sequence is processed by VGGT's track-based reconstruction pipeline, which establishes explicit cross-view correspondences and yields sparse multi-view estimates before and after bundle adjustment (BA):
\begin{equation}
\begin{aligned}
    \mathcal{H}_{\mathrm{PreBA}}
    &=\Phi_{\mathrm{track}}\!\left(\widetilde{\mathcal{X}}\right),\\
    \mathcal{H}_{\mathrm{BA}}
    &=\operatorname{BA}\!\left(\mathcal{H}_{\mathrm{PreBA}}\right).
\end{aligned}
    \label{eq:supp-vggt-backend}
\end{equation}
The two estimates share the same proxy observations and correspondence initialization, while BA additionally refines the scene structure and camera parameters by minimizing multi-view reprojection error. When this optimization is sufficiently constrained, the post-BA estimate is preferred because it provides the more globally consistent joint estimate of structure and cameras. The proxy views, however, are derived from a multiplexed SCI observation rather than directly observed images, and their correspondence support can be spatially nonuniform. If these correspondences do not sufficiently constrain the joint optimization, the refined solution may retain inadequate track support or image-plane coverage, or yield an unstable focal estimate. It can then be less reliable than the PreBA estimate, which preserves the geometry supported by the original tracks.

Accordingly, we first examine the post-BA reconstruction and retain it when refinement preserves complete view registration, a connected covisibility graph with sufficient multi-view track support, adequate image-plane point coverage, and a stable shared-focal estimate. If these conditions are not met, we revert to the corresponding PreBA reconstruction, provided that it retains adequate view registration, multi-view track support, and image-plane point coverage. Thus, PreBA is not preferred over a valid refined solution; it prevents an underconstrained joint update from replacing geometry that remains supported by the original correspondences.

When neither candidate from the initial run satisfies these criteria, we rerun the same track-based reconstruction pipeline with a larger set of 2D tracking queries under a small, fixed set of input orderings of the same $N$ proxy views. The larger query set covers more locations in the existing proxy images for cross-view tracking; it neither changes their spatial resolution nor introduces additional views. All retries use the same expanded query configuration and differ only in input ordering. Each retry again produces PreBA and post-BA candidates, to which we apply the same validity checks and preference for post-BA over PreBA. If no retry yields valid track-supported geometry, VGGT's direct camera and depth predictions provide a dense initialization. This final fallback does not require persistent cross-view tracks or the track-constrained, shared-camera BA used by the preceding pipeline. The resulting hierarchy prioritizes geometry constrained by explicit cross-view tracks and invokes the dense fallback only when such geometry cannot be recovered. All retry configurations and validity criteria are fixed across scenes and are applied without access to ground-truth imagery, ground-truth camera parameters, or final reconstruction metrics.

The retained initialization defines $\mathcal{Q}^{\star}$, $\mathbf{K}^{\star}_i$, and $\mathbf{T}^{(0)}_i$ in \MainEqref{eq:vggt}{4}. Its point cloud, camera intrinsics, and poses initialize the Gaussian representation and camera parameters. The proxy views are used only for geometric initialization and never as photometric targets during SCI optimization. Measured on the single NVIDIA RTX 4090 setup used in our experiments, the complete initialization from proxy construction to the retained VGGT geometry estimate takes 4~min 45~s per scene on average.

\section{Additional Implementation and Ablation Details}
Beyond the Opacity-Guided Splitting and Growth Regulation (OSGR) formulation in \MainSecref{sec:densification}{III-E}, we compute the local opacity reference of each Gaussian using the Gaussian itself and its three nearest neighbors. Only Gaussians with a positive deviation from this local reference enter the additional opacity-driven split set. If this set exceeds 5\% of the current Gaussian population, we retain the 5\% with the largest positive deviations. This budget applies only to the additional opacity-driven candidates: they are merged with the conventional gradient-, scale-, and screen-radius-based split candidates, and the original high-gradient duplication rule remains unchanged. The resulting rule is relative to the local opacity distribution and therefore does not require a fixed absolute opacity threshold. We also retain the transmittance-preserving opacity correction of Revising-GS~\cite{bulo2024revisingdensificationgaussiansplatting} for both duplication and splitting. In a controlled ablation on the six standard scenes in \MainTabref{tab:sci-comparison}{I}, applying this correction improves the overall reconstruction result in four cases; we therefore retain it for every scene to keep the OSGR update rule consistent across reconstructions.

The Gaussian-count constraint uses the population reached at a reference training iteration as the upper bound for subsequent splitting and duplication. This shared temporal rule adapts the numerical cap to the population accumulated by each reconstruction. Under an otherwise identical optimization protocol, we compare reference iterations of 5,000, 6,000, 7,000, 8,000, and 9,000. The 7,000-iteration setting gives the best overall result among these capped configurations and is fixed for all reported experiments. We additionally evaluate an uncapped configuration, which fails to complete the full six-scene evaluation because a subset of runs exhausts the available GPU memory before reaching the 20,000-iteration budget, as reported in \MainTabref{tab:ablation-vfm-check}{VII}.

To quantify stochastic variation, we repeat the full coarse-stage configuration in \MainTabref{tab:ablation-vfm-check}{VII} three times with different random seeds while keeping the acquisition, initialization branch, optimization schedule, and training budget unchanged. Table~\ref{tab:three_run_variance} reports the resulting mean and variance. The corresponding standard deviations are approximately 0.0435~dB for PSNR, 0.0007 for SSIM, and 0.0010 for LPIPS, indicating limited variability over the three independent runs performed with the same acquisition and optimization settings used throughout this evaluation. \MainTabref{tab:ablation-vfm-check}{VII}, \MainTabref{tab:ablation_fine_stage}{X}, and Table~\ref{tab:three_run_variance} provide complementary controlled evidence through coarse-stage component ablations, fine-stage component ablations initialized from a shared coarse reconstruction, and run-to-run variability of the complete coarse-stage configuration, respectively.

\begin{table}[!t]
\caption{Coarse-stage run-to-run stability over three independent runs. We report the three-seed mean and variance for each reconstruction metric.}
\label{tab:three_run_variance}
\centering
\small
\setlength{\tabcolsep}{10pt}
\renewcommand{\arraystretch}{1.08}
\begin{tabular}{@{}lcc@{}}
\toprule
Metric & Three-seed Mean & Variance \\
\midrule
PSNR  & 35.5689  & 0.00189259 \\
SSIM  & 0.975100 & $5.0861\!\times\!10^{-7}$ \\
LPIPS & 0.028900 & $1.0701\!\times\!10^{-6}$ \\
\bottomrule
\end{tabular}
\end{table}

\begin{figure}[!t]
    \centering
    \includegraphics[width=\linewidth]{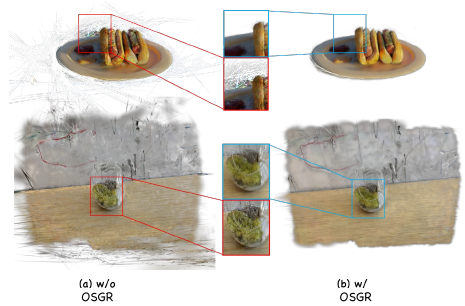}
    \caption{3D qualitative ablation of OSGR, complementing the image-space comparison in Fig.~7 of the main paper. Removing OSGR yields less coherent geometry and stronger structural distortions, whereas the full model produces a more compact and geometrically consistent representation.}
    \label{fig:ablation_gs}
\end{figure}

\begin{figure}[!t]
    \centering
    \includegraphics[width=\linewidth]{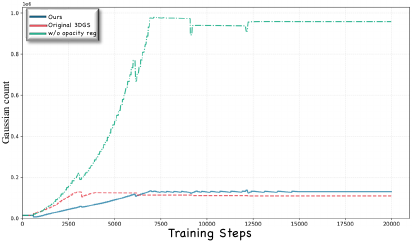}
    \caption{Gaussian-population trajectories during coarse optimization. Ours and the variant without opacity regulation use the same count-constraint protocol, which fixes the population reached at iteration 7,000 as the upper bound thereafter. Removing the soft mean-opacity term yields a substantially larger population by the reference iteration, whereas the complete model remains close to standard 3DGS adaptive density control~\cite{kerbl20233dgaussiansplattingrealtime}.}
    \label{fig:gaussian_count}
\end{figure}

\enlargethispage{\baselineskip}
Figures~\ref{fig:ablation_gs} and~\ref{fig:gaussian_count} provide two complementary representation-level diagnostics. Figure~\ref{fig:ablation_gs} is the 3D counterpart to the image-space ablation in \MainFigref{fig:ablation_2d}{7}: removing OSGR produces a less coherent Gaussian structure, whereas the full model maintains a more compact representation. Figure~\ref{fig:gaussian_count} reports the corresponding population trajectories during coarse optimization. Both OSGR configurations use the same count-constraint protocol, yet the variant without opacity regulation reaches a substantially larger population by iteration 7,000. The mean-opacity term provides soft regulation during active densification, and the population reached at the reference iteration supplies the hard upper bound thereafter. The complete configuration remains close to standard adaptive density control and substantially below the variant without opacity regulation throughout the later optimization stage.

\FloatBarrier

\section{Additional Results}
\begin{table}[!h]
\caption{Comparison of proxy-view initialization strategies over six mixed cases. Bold values indicate the best result in each column.}
\label{tab:adaptive-initialization}
\centering
\fontsize{8.3}{9.6}\selectfont
\setlength{\tabcolsep}{4.2pt}
\renewcommand{\arraystretch}{1.12}
\begin{tabular}{@{}lcccc@{}}
\toprule
Method & nATE (\%)$\downarrow$ & PSNR$\uparrow$ & SSIM$\uparrow$ & LPIPS$\downarrow$ \\
\midrule
Naive $\mathbf{Y}\odot\mathbf{M}_i$ & 17.1603 & 27.62 & 0.8240 & 0.2362 \\
ENI only & 18.3801 & 28.61 & 0.8471 & 0.2158 \\
AMDI only & 2.3468 & 31.57 & 0.8776 & 0.1629 \\
\rowcolor{highlightgray}
Adaptive (Ours) & \textbf{2.3277} & \textbf{31.72} & \textbf{0.8810} & \textbf{0.1611} \\
\bottomrule
\end{tabular}
\end{table}

\FloatBarrier

\begin{figure*}[!t]
    \centering
    \includegraphics[width=\linewidth]{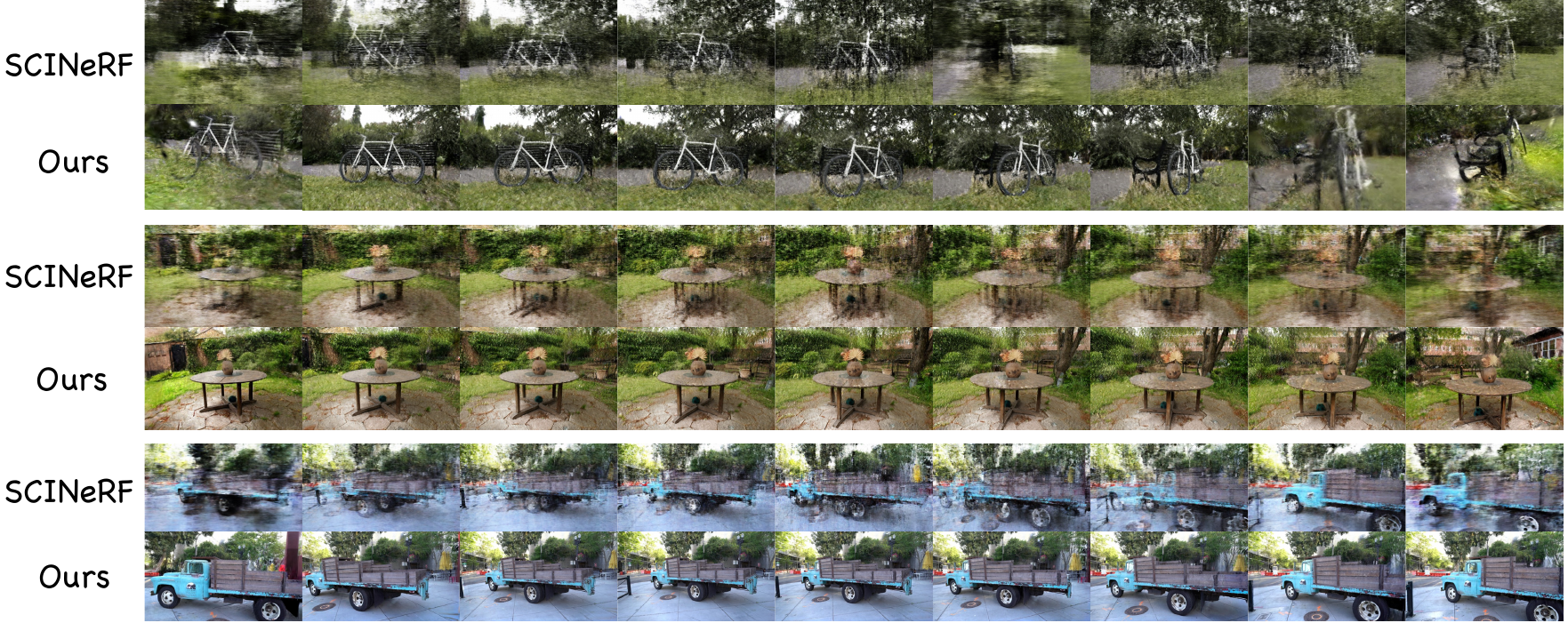}
    \caption{Additional novel-view synthesis results on Mip-NeRF 360~\cite{barron2022mip} and Tanks and Temples~\cite{knapitsch2017tanks}. Each scene group compares SCINeRF~\cite{li2024scinerfneuralradiancefields} in the upper row with our method in the lower row, while columns progress through unseen viewpoints along the rendering trajectory. The sequences provide a qualitative comparison of structural continuity and appearance consistency across the evaluated scenes and substantial viewpoint changes.}
    \label{fig:more-novel-view-mipnerf360}
\end{figure*}

\FloatBarrier

To directly evaluate proxy construction and isolate the routing decision, Table~\ref{tab:adaptive-initialization} compares a naive masked proxy, forced ENI, forced AMDI, and adaptive routing across six acquisition settings: two low-multiplicity, two high-mask-ratio, and two high-compression-ratio cases. All variants use the same VGGT backend and downstream SCI optimization, with $\tau_{\mu}=4$ and all AMDI parameters fixed. ENI is not a latent-view estimate; it forms a geometry-oriented proxy from normalized aggregate samples when overlap is limited. AMDI estimates local per-view contributions, while its smallest-admissible-window rule adapts the spatial support to local mask conditioning. Forced ENI and AMDI perform comparably in the low-multiplicity cases. Across the four stronger-mixing cases, AMDI improves over ENI by 4.40~dB PSNR, 0.0457 SSIM, and 0.0793 LPIPS on average. Adaptive routing gives the best aggregate reconstruction quality, reaching 31.72~dB PSNR, 0.8810 SSIM, and 0.1611 LPIPS, and also obtains the lowest nATE of 2.3277\%.

\begin{figure*}[!b]
    \centering
    \includegraphics[width=\linewidth]{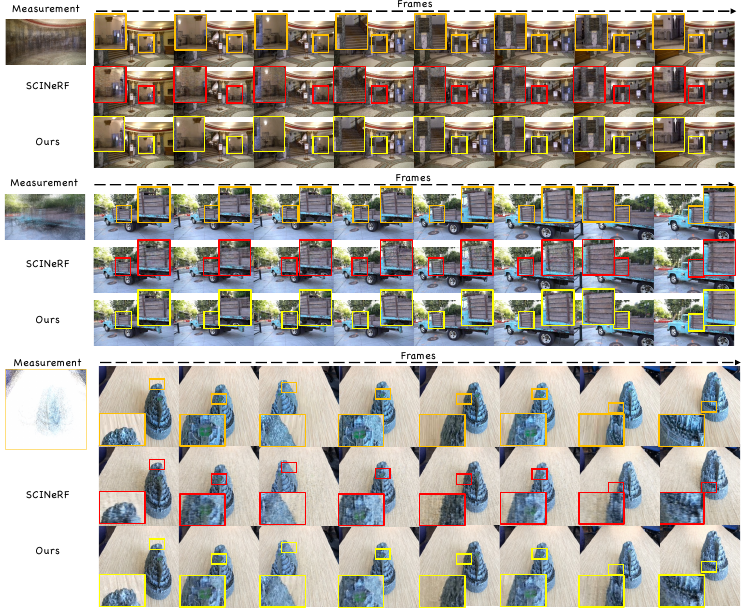}
    \caption{Additional qualitative results for scenes with complex geometry and irregular camera motion. In each scene group, the coded SCI measurement is shown on the left, followed by frames sampled along the camera trajectory. SCINeRF~\cite{li2024scinerfneuralradiancefields} and our reconstruction are presented in the upper and lower rows, respectively, enabling comparison of structural and appearance consistency across challenging viewpoint changes.}
    \label{fig:hard-case}
\end{figure*}

\begin{figure*}[!p]
    \centering
    \includegraphics[width=\linewidth]{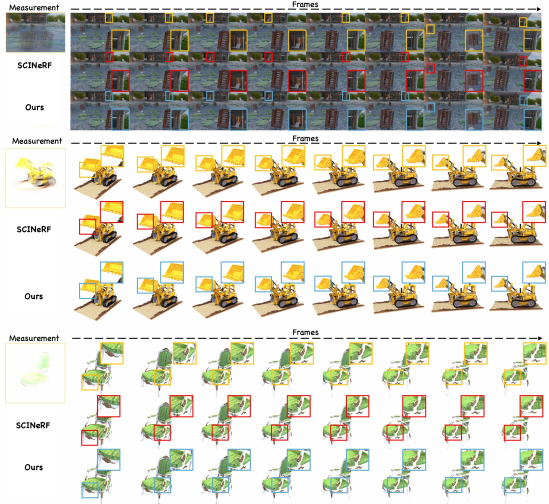}
    \caption{Additional reconstruction results on extended synthetic benchmarks. For each scene, the SCI measurement is shown on the left, followed by the reference frames and the reconstructions of SCINeRF~\cite{li2024scinerfneuralradiancefields} and our method from top to bottom. Colored boxes mark corresponding local regions for closer inspection of structural and appearance differences.}
    \label{fig:more-synthetic}
\end{figure*}

Figures~\ref{fig:more-novel-view-mipnerf360}--\ref{fig:more-synthetic} provide additional qualitative examples covering novel-view synthesis, challenging camera trajectories, and extended synthetic reconstruction. Specifically, Fig.~\ref{fig:more-novel-view-mipnerf360} compares novel-view synthesis on Mip-NeRF 360~\cite{barron2022mip} and Tanks and Temples~\cite{knapitsch2017tanks}, Fig.~\ref{fig:hard-case} examines complex geometry and irregular camera motion, and Fig.~\ref{fig:more-synthetic} presents additional synthetic-scene reconstructions.

\FloatBarrier

\end{document}